\documentclass[journal]{IEEEtran}

\newcommand{\etal}{\textit{et al}. }

\newcommand{\eg}{\textit{e}.\textit{g}. }

\usepackage{times}
\usepackage{epsfig}
\usepackage{graphicx}
\usepackage{amsmath}
\usepackage{amssymb}
\usepackage{gensymb}
\graphicspath{ {/} }
\usepackage{multirow}

\ifCLASSINFOpdf
\else
\fi
\usepackage{algorithmic}

\newsavebox{\ieeealgbox}

\usepackage{array}

\ifCLASSOPTIONcompsoc
  \usepackage[caption=false,font=normalsize,labelfont=sf,textfont=sf]{subfig}
\else
  \usepackage[caption=false,font=footnotesize]{subfig}
\fi
\begin{document}
%
\title{Homocentric Hypersphere Feature Embedding for Person Re-identification}
%
%
%

\author{Wangmeng~Xiang,
		Jianqiang~Huang,
		Xianbiao~Qi, Xiansheng~Hua,~\IEEEmembership{Fellow,~IEEE}
        and~Lei~Zhang,~\IEEEmembership{Fellow,~IEEE}

\thanks{Wangmeng Xiang, Xianbiao Qi and Lei Zhang are with the Department of Computing, The Hong Kong Polytechnic University, Kowloon, HongKong, China e-mail: (cswxiang@comp.polyu.edu.hk; csxqi@comp.polyu.edu.hk; cslzhang@comp.polyu.edu.hk).}

\thanks{Jianqiang Huang and Xiansheng Hua are with Alibaba DAMO Academy, Hangzhou, China email: (jianqiang.hjq@alibaba-inc.com; xiansheng.hxs@alibaba-inc.com).}}

\maketitle

\begin{abstract}

Person re-identification (Person ReID) is a challenging task due to the large variations in camera viewpoint, lighting, resolution, and human pose. Recently, with the advancement of deep learning technologies, the performance of Person ReID has been improved swiftly. Feature extraction and feature matching are two crucial components in the training and deployment stages of Person ReID. However, many existing Person ReID methods have measure inconsistency between the training stage and the deployment stage, and they couple magnitude and orientation information of feature vectors in feature representation. Meanwhile, traditional triplet loss methods focus on samples within a mini-batch and lack knowledge of global feature distribution. To address these issues, we propose a novel homocentric hypersphere embedding scheme to decouple magnitude and orientation information for both feature and weight vectors, and reformulate classification loss and triplet loss to their angular versions and combine them into an angular discriminative loss.
We evaluate our proposed method extensively on the widely used Person ReID benchmarks, including Market1501, CUHK03 and DukeMTMC-ReID. Our method demonstrates leading performance on all datasets.



\end{abstract}


%
\IEEEpeerreviewmaketitle

%
%
%
%


\section{Introduction}

\IEEEPARstart{P}{erson} re-identification (Person ReID) is an important computer vision task, which aims to identify a person from a set of gallery images captured under different cameras, or different timestamps under a single camera~\cite{gheissari2006person}. 
In recent years, Person ReID has drawn a lot of attentions from both academia and industry, due to its huge potential applications, such as suspect searching and multi-camera person tracking in a large-scale surveillance system. However, the task of Person ReID is extremely challenging due to the large variations in camera viewpoint, lighting, resolution, and human pose.

The key issue of Person ReID problem is how to match the query image/video with gallery images/videos. Generally speaking, both query and gallery images/videos are represented by feature vectors, which are extracted by either feature learning methods or hand crafted heuristic algorithms. This similarity of query feature vector and gallery feature vectors is then calculated by certain distance measures and the returned matching list is determined by the feature distances. Before the bloom of Convolutional Neural Networks (CNN), various heuristic representations have been used for person re-identification, such as the local maximal occurrence representation (LOMO)~\cite{7298832}, hierarchical Gaussian descriptor (GOG)~\cite{matsukawa2016hierarchical}. These representations are designed to handle light variance, pose/view changes, and so on. Other works focused on similarity/metric learning techniques, which learn robust metrics under various conditions. However, as these methods using handcraft features and metrics, they are outperformed by CNN based methods.

With the recent development of Convolutional Neural Networks (CNN)~\cite{lecun1995convolutional}, 
the performance of Person ReID has been increased dramatically. Fast evolvement in CNN architectures, such as AlexNet~\cite{alex}, VGGNet~\cite{Simonyan14c}, GoogleNet~\cite{g1} 
and ResNet~\cite{He2015}, speeds up the development of Person ReID algorithms. Meanwhile, the increasing scale of Person ReID datasets~\cite{zheng2015scalable,li2012human,li2014deepreid,ristani2016MTMC} also facilitates the study of Person ReID. Existing CNN based methods can be roughly grouped into three categories: 1) transferring and improving powerful CNN architectures to Person ReID  ~\cite{chen2017beyond,DBLP:journals/corr/SunZDW17,zhao2017deeply,varior2016gated,DBLP:journals/corr/VariorSLXW16,geng2016deep}, where off-the-shelf feature extractors are used as parts of the network architecture; 2) designing more effective metrics~\cite{DBLP:journals/corr/YiLL14,DBLP:journals/corr/ShiYZLLZL16,7298832,jose2016scalable,DBLP:journals/corr/SongXJS15,liao2015efficient}; 3) combining priori into network architecture for fine-grained feature learning~\cite{zhao2017deeply,li2017learning,zhao2017spindle,DBLP:journals/corr/ZhengZY17aa,zhang2016semantics}.

CNN with a triplet loss or classification loss is a popular framework for Person ReID. Many state-of-the-art methods~\cite{DBLP:journals/corr/HermansBL17,cheng2016person,DBLP:journals/corr/ShiYZLLZL16,li2017person,xiao2016learning} employ these loss functions or their variants during the training stage. For instance, Xiao~\etal~\cite{xiao2016learning} trained a deep CNN from scratch using datasets from multiple sources. They employed softmax loss and domain guided dropout for training and achieved competitive performance. A multi-channel part-based CNN model under the triplet framework was proposed in~\cite{cheng2016person}. This model shows superior performances on several popular benchmarks.  Hermans~\etal~\cite{DBLP:journals/corr/HermansBL17} conducted extensive experiments to validate the effectiveness of triplet loss. In~\cite{li2017person}, a two branch classification loss was employed to learn discriminative local and global features. 

Despite having achieved great successes, traditional triplet loss and classification loss have a few problems. Triplet loss focuses on optimizing local distance metrics between positive pairs and negative pairs, but lacks knowledge of global feature distribution. Classification loss overlooks the intra-class variance and only maximizes inter-class variance. Meanwhile, most current classification loss based works on Person ReID lose the measure consistency between the training stage and the deployment stage. In the training stage, these methods calculate the inner product between the features $\mathbf{x}$ and the weight vector $\mathbf{w}$. In the deployment stage, the features are first normalized. Given two $L_2$ normalized features $\mathbf{x}_a$ and $\mathbf{x}_b$, their similarity is generally calculated by cosine similarity  $ s(\mathbf{x}_a, \mathbf{x}_b) = \mathbf{x}^T_a\mathbf{x}_b = \cos\theta_{ab}$, which is determined by the angle $\theta_{ab}$ between features $\mathbf{x}_a$ and $\mathbf{x}_b$. The inconsistency mixes the orientation information and magnitude information of features, which generates a gap between the training stage and the deployment stage.

On the other hand, to achieve good retrieval performance using distance measures such as $L_2$ distance, the entries in the feature vector should better be independent. However, the weight vectors of fully-connected layers are usually correlated after the training. This leads to correlated entries in feature vectors, and causes inferior performance of learned feature representation. Several works have explored how to reduce the feature correlation in deep neural network training. DeCov~\cite{DeCov} encourages diverse representations by minimizing the cross-covariance of hidden activations. Some works reduce correlation in feature entries by imposing orthogonal constraints on weight vectors. PCANet~\cite{PCANet} which is proposed for image classification is featured by cascaded principal component analysis (PCA) filters. It uses unsupervised methods to learn orthogonal filters from raw images. In~\cite{xie2017all}, Xie~\etal applied a regularizer which utilizes orthonormality among different filter banks. SVDNet~\cite{DBLP:journals/corr/SunZDW17} achieves orthogonality by taking a restraint and relaxation iteration (RRI) training scheme, which iteratively integrates the orthogonality constraint in CNN training.

In this paper, we propose a homocentric hypersphere embedding learning  approach for Person ReID. In our framework, the weight vectors and the features are normalized to two homocentric hyperspheres with the same coordinate origin. This decouples the magnitude and orientation of feature vectors and ensures the training and testing measure consistency. Based on the homocentric hyperspheres, we jointly consider the triplet loss and softmax classification loss from the perspective of angle discrimination. In triplet loss, the optimization of distances between features is reformulated to the optimization of angular distances between features. In classification loss, the posterior probability distribution will depend solely on the angle between weight vectors and features. To ensure feature orthogonality, we add a regularization term in the loss function without relying on complicated iteration methods. The angular versions of triplet loss and classification loss work well under the unified settings and they complement each other. Meanwhile, explicitly formulating the angle between the feature vectors in training avoids the measure inconsistency between training and evaluation stages, and improves the generalization power of our approach.

We evaluate our approach on three widely used Person ReID benchmarks, including Market1501~\cite{zheng2015scalable}, CUHK03~\cite{li2014deepreid}, DukeMTMC-ReID~\cite{ristani2016MTMC}. Our method demonstrates leading results on all the evaluation datasets. Specifically, our approach achieves 78.56\% mAP and 91.28\% Rank-1 accuracy on the Market1501 dataset.

\section{Related Works}

\subsection{CNN based Person Re-Identification}

CNN based Person ReID methods consists of two key components: \emph{feature learning} and \emph{metric learning}. 

The network architecture is crucial for feature learning in Person ReID. Using pre-trained networks has been proved to be an effective strategy in many applications especially when the data scale is not big enough to train a deep network from scratch. 
Some state-of-the-art CNNs, such as AlexNet, GoogLeNet, VGGNet and ResNet, have been employed as the feature extraction module in Person ReID~\cite{chen2017beyond,DBLP:journals/corr/SunZDW17,zhao2017deeply,geng2016deep}. For example, Chen~\etal~\cite{chen2017beyond} used AlexNet as feature extractor in their deep quadruplet network. Sun~\etal~\cite{DBLP:journals/corr/SunZDW17} used CaffeNet and ResNet as backbone networks. GoogLeNet was employed as the base network in \cite{zhao2017deeply,geng2016deep}. Besides base networks, some works~\cite{zhao2017spindle,chen2017person,zhou2017large,li2017learning} developed customized modules to capture human body prior. Specifically, Zhao~\etal~\cite{zhao2017spindle} considered human body structure information in ReID pipeline. Features of different body regions are extracted separately and merged by a tree-structured fusion network. Chen~\etal~\cite{chen2017person} developed a multi-scale network architecture with a saliency-based feature fusion. Zhou~\etal~\cite{zhou2017large} built a part-based CNN to extract discriminative and stable features for body appearance.

In person Re-ID, various loss functions have been used for learning deep embedding representations, including verification loss~\cite{DBLP:journals/corr/YiLL14,ahmed2015improved,li2014deepreid,wu2016personnet}, contrastive loss~\cite{DBLP:journals/corr/VariorSLXW16}, triplet loss~\cite{DBLP:journals/corr/HermansBL17,cheng2016person,DBLP:journals/corr/ShiYZLLZL16} and quadruplet loss~\cite{chen2017beyond}.  Yi~\etal~\cite{DBLP:journals/corr/YiLL14} adopted a siamese network and softmax loss to determine whether two input images belong to the same person or not.  Shi~\etal~\cite{DBLP:journals/corr/ShiYZLLZL16} trained their network using triplet loss with hard positive pairs mining. Chen~\etal~\cite{chen2017beyond} designed a quadruplet loss with a margin based online hard negative mining. 
All these loss functions attempt to learn pairwise/ternary/quaternary distance relations.
Beyond direct metric learning, some works~\cite{xiao2016learning,li2017person,zheng2016person} address the ReID problem from the perspective of classification. Such kind of works learn an embedding metric in an indirect way by constructing clear classification boundaries. Typically, classification loss contains a softmax layer with embedding features as input. Classification loss provides global classification boundaries and metric learning constructs pairwise/ternary/quaternary distance relations.

The Person ReID problem shares similarity to the face verification and recognition problems. Recently, several works~\cite{liu2017sphereface, liu2017rethinking} in face recognition area consider hypersphere normalization constraints. In~\cite{liu2017sphereface}, SphereFace enforces the weights of the last classification layer lie on a unit hypersphere, which can be considered as a specific Weight Normalization~\cite{Salimans2016WeightNorm} with the length of weight vector as 1. 
Liu \etal~\cite{liu2017rethinking} normalized face features and optimized the distance between features and feature cluster centroids within a mini-batch. Our proposed method differs from these works by considering both feature-class center distance and feature-feature distance at the embedding learning stage. In addition, we directly learn the class center vectors, without using the mini-batch average as an approximation. Under homocentric hypersphere settings, we build a unified angular understanding for the triplet loss and the softmax classification loss. As far as we know, our work is the first to introduce normalization on both class center vectors and features and apply angular discriminative loss for Person ReID tasks.

\subsection{Orthogonal constraints on weight matrix}

The highly correlated weight vectors of fully connected layers in CNN often results in correlations among entries of feature vectors, which would reduce retrieval performance. In order to fix this problem, several works have been proposed to introduce orthogonality in neural networks~\cite{DeCov,PCANet,xie2017all,DBLP:journals/corr/SunZDW17}. Among them, SVDNet~\cite{DBLP:journals/corr/SunZDW17} uses a SVD layer as embedding layer to project high dimension features to a lower dimension manifold. The network is trained in a restraint and  relaxation iteration (RRI) scheme, which iteratively integrates the orthogonality constraint in CNN training. During the restraint stage, the weight matrix is replaced by an orthogonal matrix, which is the product of the left unitary matrix and the singular value matrix, and the rest of the network is updated with this matrix fixed. At the relaxation stage, the restraint is removed and the whole network is tuned end-to-end. This procedure is repeated several times until the convergence criteria are met. The SVD layer can be learnt by back propagation. However, the training procedures are quite complicated and the iterative training could take a long time. Our method differs from previous methods by directly adding an orthogonal regularization term to the loss function, which is  structurally simple, effective, and efficient in training.


\section{Homocentric Hypersphere Embedding}

\subsection{Triplet Loss and Softmax Loss}

We briefly discuss triplet loss and softmax loss to draw forth the proposed angular version of these loss functions under the homocentric hypersphere assumption.

\textbf{Triplet Loss.} Triplet loss was firstly proposed in FaceNet~\cite{schroff2015facenet} to improve face recognition and verification. It is originally derived from the Large Margin Nearest Neighbor (LMNN) method~\cite{weinberger2006distance}.
The objective of triplet loss is to learn an embedding function $f:\mathbf{I \mapsto x}$ so that the embedded features of images from the same class are closer than the embedded features of images from different classes in the embedding space. A sample triplet $(\mathbf{I}_a, \mathbf{I}_p, \mathbf{I}_n)$ consists of an anchor sample image $\mathbf{I}_a$, a positive sample $\mathbf{I}_p$ and a negative sample $\mathbf{I}_n$.  $(\mathbf{x}_a,\mathbf{x}_p,\mathbf{x}_n)$ are the corresponding embedded features of $(\mathbf{I}_a, \mathbf{I}_p, \mathbf{I}_n)$. The general formulation for the triplet loss can be represented as follows:

\begin{equation}
 \mathcal{L}_{t} = \sum_{i \in B} \left[ \underbrace{\left  \|\mathbf{x}_a^i-\mathbf{x}_p^i\right \|_{2}^{2}}_{pull}  \overbrace{- \left \| \mathbf{x}_a^i-\mathbf{x}_n^i\right \|_{2}^{2}}^{push} + m \right ]_+,
\label{eq:triplet}
\end{equation}

\noindent where $[\sigma]_+$ denotes $\max(\sigma,0)$, $m$ is a preset margin, and $B$ is the number of triplets.

Eq. \ref{eq:triplet} contains a pull term to pull samples from the same class together, and a push term to push samples from different classes away.
The training process of the neural network is to optimize the embedding function $f$ (adjust network parameters) to ensure that, given the feature $\mathbf{x}_a^i$ of an anchor sample $\mathbf{I}_a$ in the $i$-th triplet, the distance between $\mathbf{x}_p^i$ and $\mathbf{x}_a^i$ is smaller than the distance between $\mathbf{x}_n^i$ and $\mathbf{x}_a^i$ by a margin $m$.

One key issue in training Deep CNN with the triplet loss is hard triplet mining. The learning process may be dominated by simple triplets, and thus fail to learn an effective embedding. In this work, we employ an online hard triplet mining strategy in which we only consider the hardest triplets within a mini-batch. During the training phase, for each mini-batch, we randomly select $P$ identities, and for each identity, we randomly select $N$ samples. Therefore, each mini-batch consists of $P \times N $ samples. For each sample, we construct its hard triplet by choosing its farthest positive sample feature as $\mathbf{x}_p$ and its closest negative sample feature as $\mathbf{x}_n$. In this way, we modify the triplet loss with the online hard triplet mining as
\begin{equation}
 \mathcal{L}_{ht} = \sum_{i \in P\times N} \left[ \left \|\mathbf{x}_a^i-\mathbf{x}_{fp}^i\right \|_{2}^{2}  - \left \| \mathbf{x}_a^i-\mathbf{x}_{cn}^i\right \|_{2}^{2} + m \right ]_+, 
\label{eq:triplet-hard}
\end{equation}

\noindent where $\mathbf{x}_{fp}^i$ denotes the farthest one among the $N-1$ positive sample features, and $\mathbf{x}_{cn}^i$ stands for the closest one among the $(P-1)\times N$ negative sample features for the anchor $\mathbf{x}_a^i$.

From Eq.~\ref{eq:triplet-hard}, we can see that the effect of triplet loss with online hard example mining is largely affected by the mini-batch size. With a larger batch size and more samples for each identity, it is more likely to find a farther positive sample and closer negative sample, resulting in a harder triplet. It is also worth noticing that triplet loss is effective in learning pairwise/ternary distance relations; however, it only considers samples within a mini-batch without considering global feature distribution. The difficulty of training deep CNN with triplet loss will increase with the growth of dataset scale, because the number of triplets will increase exponentially with the increase of the number of training samples.

\textbf{Softmax Loss.} Softmax loss~\cite{bishop2006pattern} converts an input feature into a posterior probability distribution. In softmax loss, the predicted posterior probability for the $\ell$-th class is calculated as follows:

\begin{equation}
 p_{\ell} = \frac{\exp(z_{\ell})}{\sum_{k=1}^K{\exp(z_k)}},
\label{eq:softmax}
\end{equation}

\noindent with

\begin{equation}
 z_{\ell} = \mathbf{w}_{\ell}^T \mathbf{x} + b_{\ell},
\label{eq:softmax2}
\end{equation}

\noindent where $\mathbf{w_{\ell}}$ and $b_{\ell}$ are weight vector and bias of the last fully-connected layer for the $\ell$-th class, $K$ is the total number of classes.

With the posterior probability distribution of each input feature, a cross-entropy loss can be calculated as follows:
\begin{equation}
 \mathcal{L}_{c} = \sum_{i}{-\log(p_{y_i})},
\label{eq:crossentropysoftmax}
\end{equation}
\noindent where $y_i$ is the label of the $i$-th input sample.

\begin{figure}[t]
\includegraphics[scale=0.175]{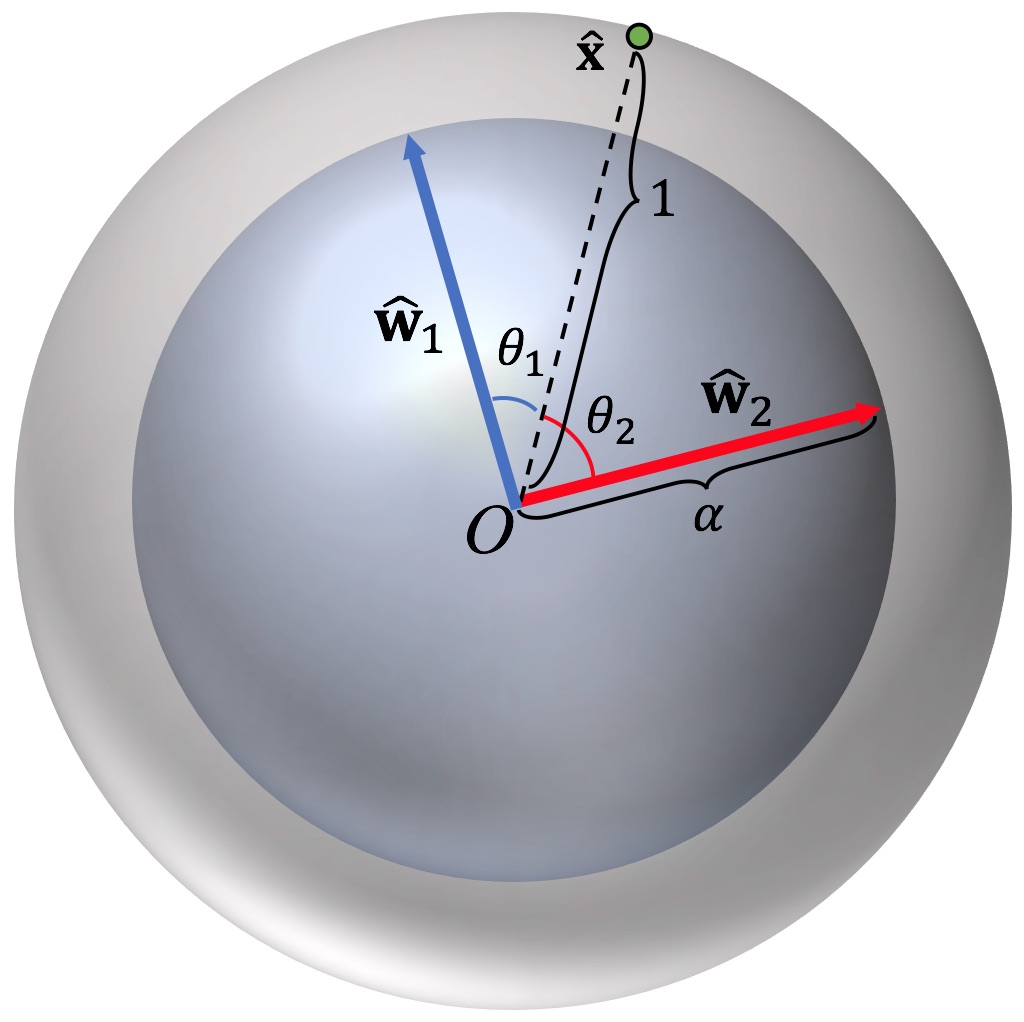}
\centering
\caption{Illustration of two homocentric hyperspheres. The small blue hypersphere is the weight hypersphere and the gray large hypersphere is the feature hypersphere. Weight vectors $\mathbf{\hat{w}}_{\ell}$ are represented by arrow lines with magnitude $\alpha$ and different colors denote for different classes. Feature vectors $\mathbf{\hat{x}}$ lie on the outer hypersphere with unit radius. $\theta_1$ and $\theta_2$ are angle distances between the feature $\mathbf{\hat{x}}$ and weight vectors $\mathbf{\hat{w}}_1$ and $\mathbf{\hat{w}}_2$, respectively.}
\label{fig:hypersphere}
\end{figure}

As indicated in the introduction section, in the deployment stage, the angle distances between the features themselves, and between the features and the weight vectors are crucial. However, the original triplet loss and softmax loss do not explicitly take the angle distance into account. 
There is no previous work targeting to address this problem in Person ReID. To overcome this issue, we propose a novel homocentric hypersphere feature embedding learning method, as described in the next subsection.

\subsection{\ Homocentric Hypersphere Constrained Embedding Learning}

\textbf{Homocentric Hypersphere.} The angle distances between features, and between  features $\mathbf{x}$ and weight vectors $\mathbf{w}_{\ell}$ are discriminative. To ensure the features $\mathbf{x}$ and the weight vectors $\mathbf{w}_{\ell}$ stay in a unified coordinate space, we propose a homocentric hypersphere feature embedding scheme. In our homocentric hypersphere, the weight vectors $\mathbf{w}_{\ell}$ and the features $\mathbf{x}$ lie on two individual hyperspheres but with the same origin. The proposed homocentric hypersphere is defined as 

\begin{equation}
\label{eq:normalization}
    \mathbf{\hat{w}}_{\ell}=\alpha \frac{\mathbf{w}_{\ell}}{\left \| \mathbf{w}_{\ell} \right \|}, \mathbf{\hat{x}} = \frac{\mathbf{x}}{\left \| \mathbf{x} \right \|},
\end{equation}

\noindent{where $\mathbf{\hat{w}}_{\ell}$ is an $L_2$ normalized weight vector scaled by a factor $\alpha$, and $\mathbf{\hat{x}}$ is an $L_2$ normalized unit feature vector.}

As shown in Fig. \ref{fig:hypersphere}, the weight vectors $\mathbf{\hat{w}_{\ell}}$ lie on a hypersphere with radius $\alpha$, and the feature $\mathbf{\hat{x}}$ lies on a hypersphere with unit radius. These two hyperspheres have the same origin $\mathbf{o}$. In the hypersphere space, closer features have a smaller angle distance. A feature will be more likely assigned to the category whose weight vector has a smaller angle with the feature. As shown in Fig. \ref{fig:hypersphere}, the feature $\mathbf{\hat{x}}$ will be more likely assigned to class 1 since it has a smaller angle distance ${\theta}_1$ to $\mathbf{\hat{w}}_{1}$.

Based on the proposed homocentric hypersphere, we reformulate the traditional triplet loss and classification loss into a new angular triplet loss and a new angular classification loss, respectively. Finally, we consider these two new losses in a unified angular framework.

\begin{figure}[t]
\includegraphics[scale=0.165]{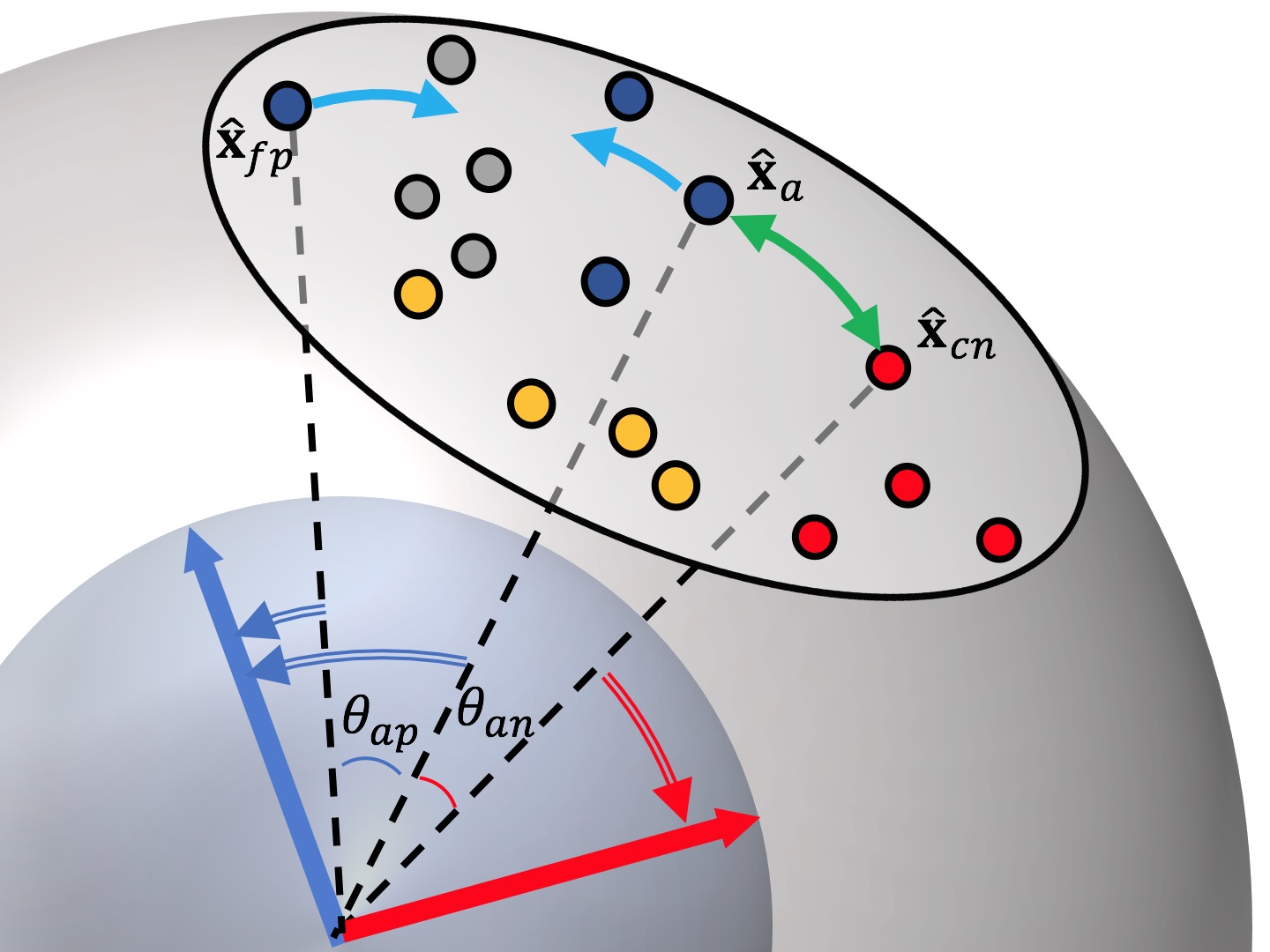}
\centering
\caption{The illustration of joint angular loss. The small circles within gray ellipse are features extracted from the same mini-batch. Different colors stand for different labels and all the features lie on the same feature hypersphere. Classification centers are represented by blue and red arrow lines. The green arrows show the pushing term of the triplet, while light blue arrows show the pulling term. The blue and red arrows with double lines demonstrate the guidance effect of class center vectors.}
\label{fig:triplet_sphere}
\end{figure}

\textbf{Angular Triplet Loss.} Distance measure on triplet loss is essential for feature learning. In the homocentric hypersphere space, the features are normalized unit vectors. Given  features $\mathbf{\hat{x}}_1$ and $\mathbf{\hat{x}}_2$, $\mathbf{\hat{x}}_1-\mathbf{\hat{x}}_2$ is a chord on feature hypersphere. As three edges of a triangle $(\mathbf{\hat{x}}_1,\mathbf{\hat{x}}_2,\mathbf{\hat{x}}_1-\mathbf{\hat{x}}_2)$ are known, the angle between features $(\mathbf{\hat{x}}_1,\mathbf{\hat{x}}_2)$ can be represented as

\begin{equation}
    \theta = 2 \arcsin(\frac{\left \| \mathbf{\hat{x}}_1-\mathbf{\hat{x}}_2 \right \|_{2}^2}{2}),
\label{eq:angledistance}
\end{equation}

\noindent where $\theta \in [0, \pi]$. As shown in Fig.~\ref{fig:triplet_sphere}, given the triangle $(\mathbf{\hat{x}}_a,\mathbf{\hat{x}}_{cn},\mathbf{\hat{x}}_a-\mathbf{\hat{x}}_{cn})$, the angle $\theta_{an}$ between $\mathbf{\hat{x}}_a$ and $\mathbf{\hat{x}}_{cn}$ can be calculated using Eq.~\ref{eq:angledistance}.

Under the definition of homocentric hypersphere, it is more natural to measure the angle distance between features rather than Euclidean distance. Therefore, we reformulate Eq. \ref{eq:triplet-hard} into a new angular triplet loss as 

\begin{equation} 
 \mathcal{L}_{at} = \sum_{i \in P\times N} \left[ \theta_{ap}^i- \theta_{an}^i + \theta_m \right ]_+,
\label{eq:angulartriplet}
\end{equation}

\noindent where $\theta_{ap}^i$ stands for the angle between the anchor feature and the hardest (farthest) positive feature. $\theta_{an}^i$ represents the angle between the anchor feature and the hardest (closest) negative feature. $\theta_m$ is an angular margin.

As illustrated in Fig.~\ref{fig:triplet_sphere}, we can see that the role of push term in Eq.~\ref{eq:triplet} is now to enlarge $\theta_{an}$ and the role of pull term in Eq.~\ref{eq:triplet} is to minimize $\theta_{ap}$ w.r.t. angular margin $\theta_m$.

\textbf{Angular Classification Loss.} In the homocentric hypersphere formulation, the features and the class center vectors are enforced to have the same origin $\mathbf{o}$. To ensure the requirement, we set the bias term $b$ in the original softmax loss function to 0. In this way, we can reformulate Eq.~\ref{eq:softmax2} as

\begin{equation} 
 z_{\ell} = \alpha\frac{\mathbf{w}_{\ell}^T \mathbf{x}}{\left \|\mathbf{w_{\ell}}\right \|\left \|\mathbf{x}\right \|} = \alpha \cos \theta_{\ell},
\label{eq:angleclass}
\end{equation}
\noindent where $\theta_{\ell}$ is the angle between $\mathbf{w}_{\ell}$ and $\mathbf{x}$. In this case, the original softmax loss function in Eq. \ref{eq:crossentropysoftmax} can be reformulated as
\begin{equation}
 \mathcal{L}_{ac} = \sum_{i}{-\log(\frac{\exp(\alpha  \cos \theta_{y_i})}{\sum_{k=1}^K{\exp(\alpha\cos \theta_k)}})}.
\label{eq:anglesoftmax}
\end{equation}

 According to Eq.~\ref{eq:angleclass} and Eq.~\ref{eq:anglesoftmax}, in the training stage, our target is to maximize the cosine similarity between the features and the weight vectors of their corresponding class. This is actually minimizing the angle distance between them. The weight vector $\mathbf{w_{\ell}}$ will accumulate information from all the samples for training, and can be regarded as the class center vector of class $\ell$. The posterior probability $p_{\ell}$ for class $\ell$ now depends solely on the angle $\theta_{y_i}$ between feature vector $\mathbf{x}$ and class center vector $\mathbf{w}_{\ell}$.
 In the testing stage, the input embedding feature will rank its neighbors according to the angle distances. Therefore, the similarity measures in the training stage and the testing stage are consistent in nature.

\textbf{Joint Angular Loss.} In the previous development, we have reformulated the triplet loss and classification loss into their angular versions. Here, we combine these two loss functions as 

\begin{equation}
 \mathcal{L}_a = \mathcal{L}_{at} + \lambda\mathcal{L}_{ac},
\label{eq:jointloss}
\end{equation}

\noindent where $\mathcal{L}_{at}$ and $\mathcal{L}_{ac}$ stand for angular triplet loss and angular classification loss, respectively. $\lambda$ is a trade-off parameter between these two loss functions.

The proposed joint angular loss in Eq.~\ref{eq:jointloss} is a natural way to combine triplet loss and softmax loss under a unified angular framework. $\mathcal{L}_{ac}$ draws clear classification boundaries between categories, and $\mathcal{L}_{at}$ gives a specific pairwise/ternary angle relation.
The angular margin $\theta_m$ in $\mathcal{L}_{at}$ has a direct connection to feature discrimination on the person feature manifold, while the identity classification center distribution in $\mathcal{L}_{ac}$ provides extra information to guide feature embedding learning. 

In Fig.~\ref{fig:triplet_sphere}, we illustrate how joint angular loss works. We view the optimization of the joint angular loss from an angular discrimination perspective. For each mini-batch, we construct hard triplet by selecting the hardest positive feature $\mathbf{\hat{x}}_{fp}$ with the longest angle distance to anchor feature $\mathbf{\hat{x}}_a$, and the hardest negative feature $\mathbf{\hat{x}}_{cn}$ with the smallest angle distance to $\mathbf{\hat{x}}_a$. The angular triplet loss maximizes $\theta_{an}$ and minimizes $\theta_{ap}$ w.r.t. to angular margin $\theta_m$. The angular classification loss minimizes the angles $\theta_{y_i}$ between features and their corresponding class center vectors. The class center vectors from angular classification loss guide the feature embedding learning. This accelerates model training by preventing optimization from getting stuck in local metric learning.

\subsection{Orthogonal  Constraints  on  Embedding  Layer}
The backbone CNN is often followed by a linear embedding layer to project high dimensional features into a low dimensional feature space. The correlation among weight vectors in the embedding layer can compromise the performance of descriptors . It is important to reduce the redundancy of weight vectors in the embedding layer. Many works~\cite{DeCov,PCANet,xie2017all,DBLP:journals/corr/SunZDW17} show that an orthogonal constraints on the weight matrix is effective to achieve this goal. This motivates us to place orthogonal constraints on the embedding layer by forcing  $\mathbf{W}_e^T \mathbf{W}_e$ to be a diagonal matrix, where $\mathbf{W}_e$ represents the weight of embedding layer. Specifically, we define

 \begin{equation}
       R_e = \sum || \mathbf{W}^T_e \mathbf{W}_e - \mathbf{I} ||
 \label{eq:embeddingconstraints}
 \end{equation}
 
\noindent and add this regularization term into the loss function. The whole loss function now becomes

 \begin{equation}
    \mathcal{L} = \mathcal{L}_{a} + \gamma R_{e},
\label{eq:totalloss}
 \end{equation}
 
\noindent where $\gamma$ is a trade-off parameter.

We adopt the metric used in \cite{DBLP:journals/corr/SunZDW17} to measure the orthogonality of embedding weights: 

 \begin{equation}
    S(\mathbf{W}_e) = \frac{\sum^{k}_{i=1}{g_{ii}}}{\sum^{k}_{i=1}{\sum^{k}_{j=1}{|g_{ij}|}}},
\label{eq:sw}
 \end{equation}
 
\noindent where $g_{ij}$ denotes the entries in $\mathbf{W}^{T}_{e}\mathbf{W}_{e}$ and $k$ represents the number of weight vectors in $\mathbf{W}_e$. The values of $S(\mathbf{W}_e)$ range within $[\frac{1}{k},1]$. A higher value of $S(\mathbf{W}_e)$ indicates better orthogonality.

We conduct cross validation to find how the embedding layer's orthogonality affect the performance of trained models, and determine the parameter $\gamma$. More details can be found in section IV.D.

\subsection{The Flowchart of Algorithm}

\begin{figure}[t]
\includegraphics[scale=0.22]{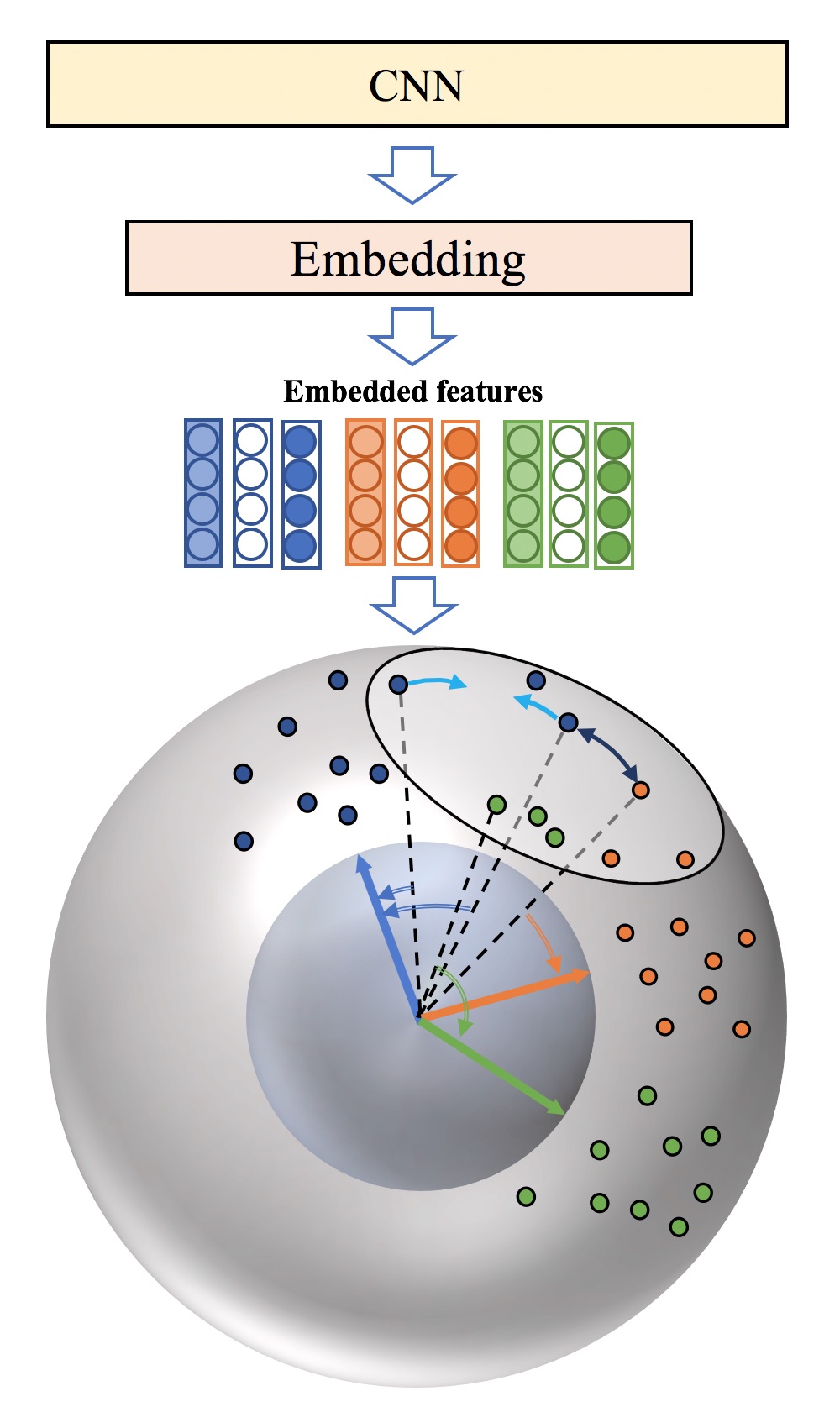}
\centering
\caption{ The flow chart of our proposed algorithm. The top part is a CNN for feature extration and the middle part is the embedding layer. The bottom illustrates our proposed homocentric hypersphere embedding learning part. The different colors of embedded features represent different identities. The embedded features are represented by the points on the feature hypersphere.}
\label{fig:framework}
\end{figure}

 The flowchart of our algorithm is shown in Fig.~\ref{fig:framework}. We divide our model into two parts: feature extraction part and metric learning part. For feature extraction, different CNN architectures can be employed, such as SqueezeNet, VGG and ResNet. We use ResNet50 as the default feature extraction network. All the CNNs are pre-trained on the ImageNet~\cite{deng2009imagenet} dataset. We apply average pooling on the final convolutional feature maps to obtain feature vectors. For the metric learning strategy, we employ an embedding layer after average pooling to generate low-dimensional features. The homocentric hypersphere constraints are then applied on features and class center vectors. Joint angular loss is employed for metric learning with additional orthogonal regularization.

\section{Experiments}

\subsection{Datasets}
We evaluate our approach on three widely used Person ReID benchmarks, including Market1501~\cite{zheng2015scalable}, CUHK03~\cite{li2014deepreid} and DukeMTMC-ReID~\cite{ristani2016MTMC}.

\textbf{Market1501} contains 32,668 pedestrian images of 1,501 identities. The images were collected by five high-resolution and one low-resolution cameras. The resolution of pedestrian images is $64 \times 128$. Each identity in the dataset appears on at least two different views. Following the official split of training and testing sets, we use 751 identities for training, and the rest 750 for testing.

\textbf{CUHK03} has 1,467 identities from two different views in the CUHK campus. There are 14,097 images obtained by Deformable Part based Models (DPM)~\cite{felzenszwalb2010object}, and 14,096 images obtained by manually labeling. 1,367 person identities are used for training, and the rest 100 identities for testing. In this paper, we follow the official evaluation protocols and report performance of our method using both DPM detected images and manually labeled images.

\textbf{DukeMTMC-ReID} is a subset of the multi-target, multi-camera pedestrian tracking dataset. In this work, we use a subset of~\cite{ristani2016MTMC} provided by~\cite{zheng2017unlabeled}, which has 16,522 images from 702 identities for training. The images were captured by eight different high-resolution cameras. The images of pedestrians are manually cropped. For testing, it has 2,228 images for querying and 17,661 gallery images from 702 identities. We follow the evaluation protocol in~\cite{zheng2017unlabeled}.

\begin{figure}
	\centering
    \subfloat[Market1501]{
        \includegraphics[width=0.45\textwidth]{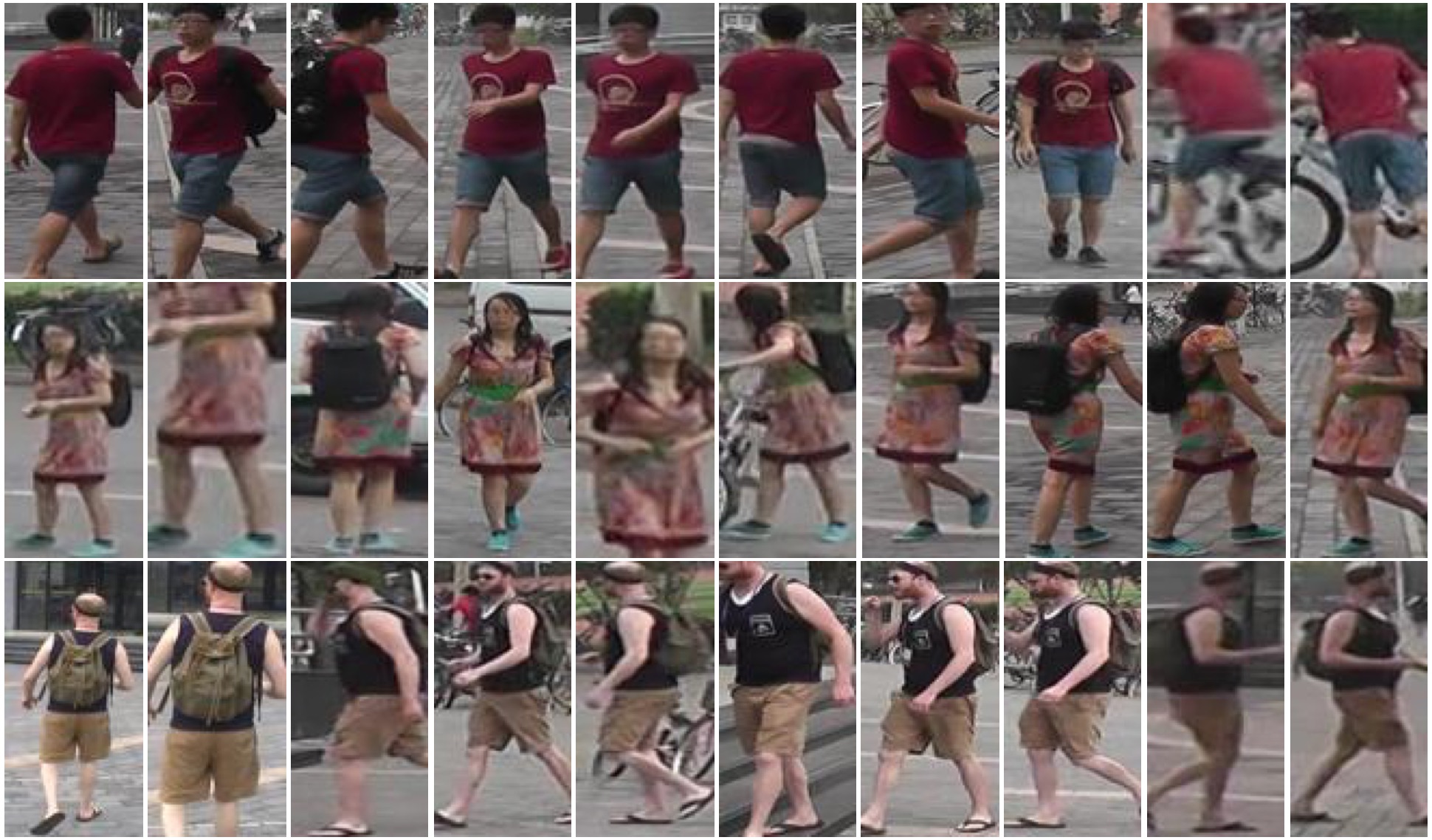}
    }

	\subfloat[CUHK03]{
        \includegraphics[width=0.45\textwidth]{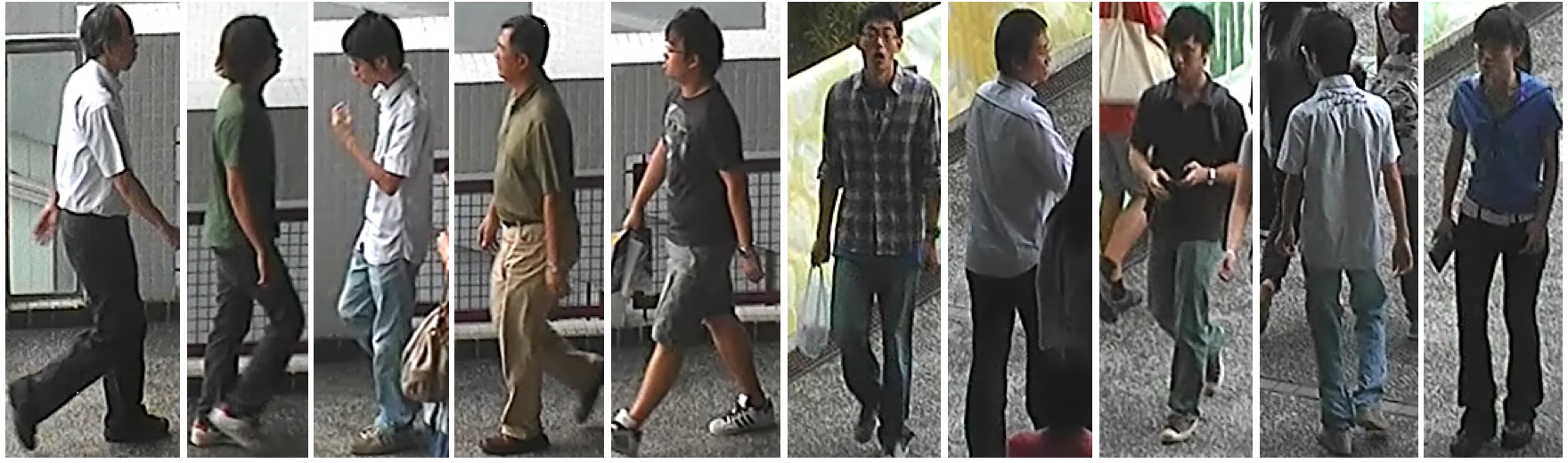}    
    }
    
	\subfloat[DukeMTMC]{
        \includegraphics[width=0.45\textwidth]{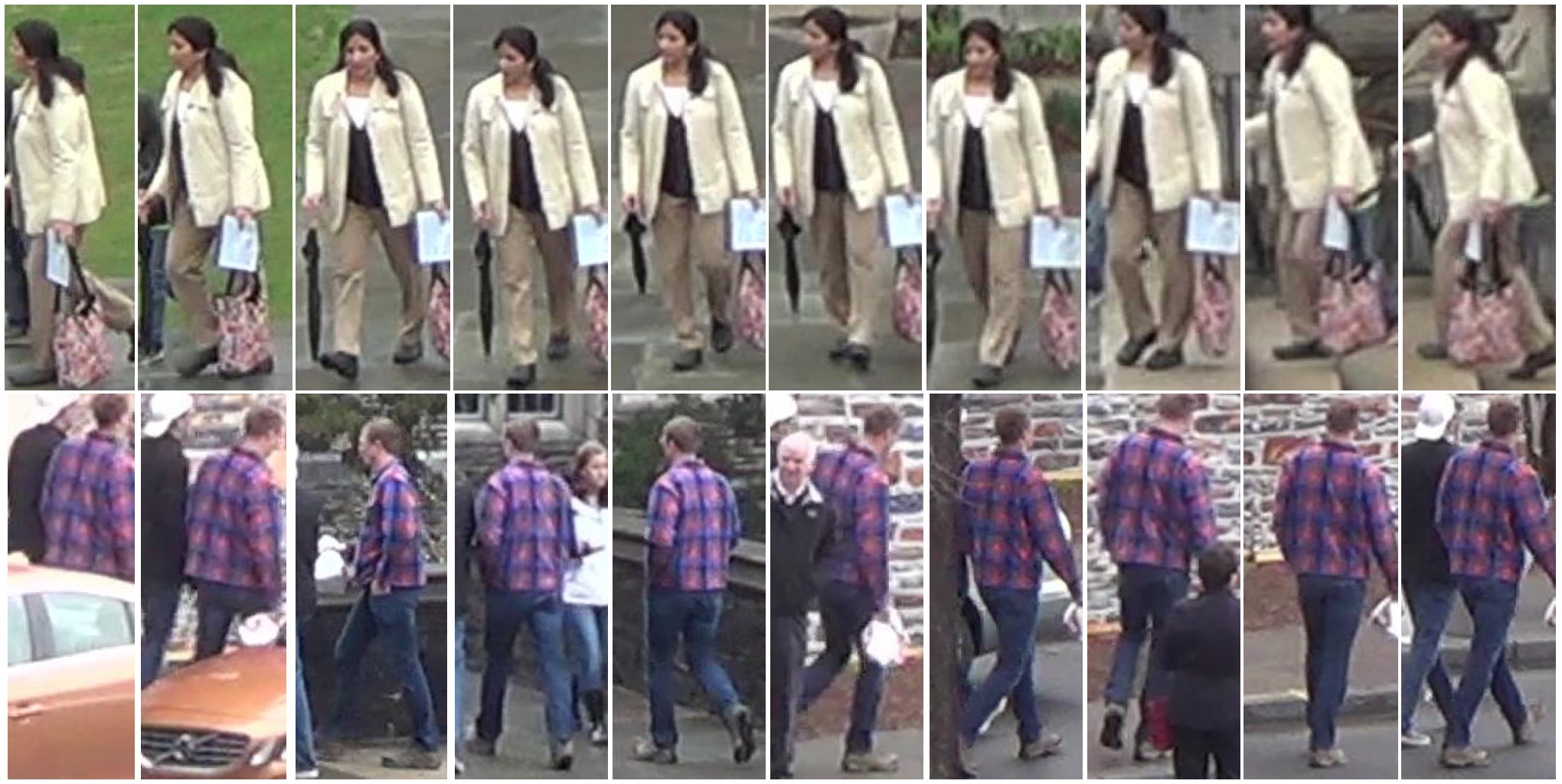}
    }

    \caption{Sample images from Market1501, CUHK03 and DukeMTMC datasets.}\label{fig:data}
\end{figure}

Table~\ref{table:data} provides a statistical summary of each dataset. It lists the number of identities (ID), bounding boxes (BBoxes), distractors (Distra), and views (Views) in each dataset. Fig.~\ref{fig:data} shows some sample images for each dataset.

\begin{table}\normalsize
\begin{center}
\caption{ The Statistics of Person ReID Datasets.}
\label{table:data}
\begin{tabular}{|l|l|c|c|c|c|}
\hline
 Datasets & ID & BBoxes & Distra & Views\\
\hline\hline
 Market1501  & 1501 & 32668 & 2793 & 6\\
 CUHK03 & 1467 & 14097 & 0 & 2 \\
 DukeMTMC-ReID & 702 & 16522 & 0 & 8 \\
\hline
\end{tabular}
\end{center}
\end{table}

\subsection{Evaluation Protocol}

We use Single-Query (SQ) by default for all datasets. According to previous evaluation protocols used in each dataset, we adopt three evaluation protocols, including Cumulated Matching Characteristics (CMC) (top1, top5 and top10), mean Average Precision (mAP), and Rank-1 identification rate. On Market1501 and DukeMTMC-ReID, query and gallery sets may have the same camera views, but for each individual query identity, his/her gallery samples from the same cameras are excluded. On CUHK03, we follow the standard testing protocol. For each query, we randomly sample one instance for each gallery identity, and compute a CMC curve in the single-gallery-shot setting. As random selection is involved, we repeat the evaluation procedure for 10 times and report the mean results. On Market1501, we report CMC and mAP. On CUHK03, we report CMC for both detected and labeled datasets. On DukeMTMC-ReID, we report Rank-1 identification rate and mAP.

\subsection{Implementation Details}

\textbf{Data preprocessing.} We use the same data pre-processing methods on all datasets. In the training stage, common data augmentation methods are applied to images, including random flipping, shifting, zooming, cropping and random erasing~\cite{randomerase}. The images are then resized to $256 \times 128$. As we use pre-trained networks from torchvision\footnote{https://github.com/pytorch/vision}, all images' pixel values are normalized to $[0,1]$, subtracted by mean pixel values of RGB channels and then divided by standard deviation of each channel.

\textbf{Optimization.} For joint angular loss, our model is trained with the batch size equals 256, and for each instance we randomly select 8 samples. We apply Adam optimizer and set the original learning rate to $1 \times 10^{-3}$ for the first 50 epochs and gradually decrease it to $1 \times 10^{-4}$ for next 50 epochs and $10^{-5}$ for the last 50 epochs.
For homocentric hypersphere embedding learning with orthogonal constraints, extensive experiments are conducted to illustrate the effect of hyper parameters of the model. The detailed analysis and discussion on parameter setting would be given in the next subsection.
 We use the same training schedule for triplet loss baseline. For classification loss baseline, we apply SGD momentum with Nesterov. We set initial learning rate to 0.1 for metric learning layers and 0.01 for pre-trained feature extraction  layer. We decrease the learning rate by a factor of 10 for every 40 epochs until  convergence. Our implementation is built on modified Open-ReID\footnote{https://cysu.github.io/open-reid/}, an open source Person ReID library.

In the following subsections, we abbreviate our method with the proposed Joint Angular Loss as JAL.

\subsection{Exploratory Experiments}

In this section, we conduct experiments to determine the values of hyperparameters in our model, including the trade-off parameter $\lambda$ between angular triplet loss and classification loss, angular margin $\theta_m$, scaling parameters $\alpha$ and orthogonal contraint parameter $\gamma$. We further show the influences of different loss functions, orthogonal constraints and test data augmentation to the ReID performance.

 \textbf{Choice of $\lambda$ and angular margin $\theta_{m}$.} As we can see from Eq.~\ref{eq:angulartriplet} and Eq.~\ref{eq:jointloss}, $\lambda$ controls the trade-off between angular triplet loss and classification loss, while angular margin $\theta_m$ controls the boundary distance of identities on feature embedding manifold. We cross validate $\lambda$ and $\theta_m$ on the ReID results on Market1501. The value of $\lambda$ is selected from $\{0.1, 0.2, 0.5, 1\}$ and $\theta_m$ is selected from $\{1^{\circ}, 3^{\circ}, 5^{\circ}, 8^{\circ}, 10^{\circ}\}$. We experimentally found that models with reasonably higher weight of triplet loss and smaller angular margin tend to have better performance on validation set. As we can see from Tab.~\ref{table:lambda}, the best $\lambda$ and $\theta_m$ on the Market1501 are around $0.2$ and $3^{\circ}$, respectively. The Table also shows that our model is insensitive to these two parameters. To simplify the experiment, we set $\lambda = 0.2$ and $\theta_m = 3^{\circ}$ in all the later experiments.

\begin{table}\normalsize
\begin{center}
\caption{Influences of different $\lambda$ and $\theta_m$ on model performance on validation set of Market1501.}
\label{table:lambda}
\begin{tabular}{|l|l|l|l|c|c|c|c|}
\hline
$\lambda$ & $\theta_m$ & Top1 & Top5 & Top10 & mAP\\
\hline\hline
 0.1 & 1 & 89.70 & 96.47 & 97.95 & 75.18\\
 0.1 & 3 & 90.11 & 96.73 & 98.01 & 76.10\\
 0.1 & 5 & 89.82 & 96.59 & 97.98 & 76.16\\
 0.1 & 8 & 90.05 & 96.29 & 97.45 & 75.96\\
 0.1 & 10 & 89.79 & 96.47 & 97.95 & 76.22\\
0.2 & 1 & 91.11 & 96.52 & 98.00 & 77.88\\
\textbf{0.2} & \textbf{3} & \textbf{90.95} & \textbf{96.78} & \textbf{97.94} & \textbf{78.05}\\
0.2 & 5 & 91.01 & 96.51 & 97.92 & 77.77\\
0.2 & 8 & 90.53 & 96.38 & 97.73 & 77.31\\
 0.2 & 10 & 90.65 & 96.14 & 97.63 & 77.13\\
 0.5 & 1 & 90.83 & 96.59 & 97.71 & 77.36\\
 0.5 & 3 & 90.97 & 96.50 & 97.92 & 77.76\\
 0.5 & 5 & 91.06 & 97.12 & 98.04 & 77.81\\
 0.5 & 8 & 90.56 & 96.44 & 97.83 & 77.78\\
 0.5 & 10 & 90.47 & 96.18 & 97.77 & 76.40\\
 1.0 & 1 & 90.38 & 96.23 & 97.77 & 76.48\\
 1.0 & 3 & 90.68 & 96.53 & 97.92 & 77.22\\
 1.0 & 5 & 90.62 & 96.35 & 97.71 & 76.08\\
 1.0 & 8 & 89.55 & 95.87 & 97.71 & 75.23\\
 1.0 & 10 & 90.08 & 96.02 & 97.51 & 76.82\\

\hline
\end{tabular}
\end{center}

\end{table}

\begin{table}\normalsize
\begin{center}
\caption{Influences of different $\alpha$ on model performance on market1501.}
\label{table:alpha12}
\begin{tabular}{|l|c|c|c|c|}
\hline
$\alpha$ & Top1 & Top5 & Top10 & mAP\\
\hline\hline
3 & 85.81 & 93.68 & 96.02 & 69.20\\
6 & 87.00 & 94.54 & 96.56 & 73.89\\
9 & 89.80 & 95.84 & 97.28 & 76.04\\

\textbf{12} & \textbf{90.95} & \textbf{96.78} & \textbf{97.94} & \textbf{78.05}\\
15 & 90.94 & 96.94 & 97.88 & 77.24\\
18 & 90.52 & 96.82 & 98.03 & 76.96\\
20 & 90.80 & 96.74 & 98.10 & 76.53 \\
30 & 89.64 & 96.37 & 97.69 & 75.36 \\
\hline
\end{tabular}
\end{center}

\end{table}

\textbf{Feature and weights scaling.} According to Eq.~\ref{eq:anglesoftmax},  there is a hyper parameters $\alpha$ to adjust the scale of the inputs to the angular softmax layer after the feature vectors and weights normalization. We conducted experiments on different $\alpha$, to study how the scaling parameter affect the performance. As we apply angular triplet loss, the value of $\alpha$ would not affect the triplet part. We select the value of  $\alpha$ from $\{3, 6, 9, 12, 15, 18, 20, 30\}$. As it is showed in Tab.~\ref{table:alpha12}, the model performance is much influenced by the value of $\alpha$. However, the trained models generally work well in a range of $\alpha$. For simplicity, we keep $\alpha = 12$ in all our experiments, and our proposed method works very robustly.

\begin{table}\normalsize
\begin{center}
\caption{Influences of different $\gamma$ on model performance on Market1501. SW is the measure of orthogonality of weights.}
\label{table:gamma}
\begin{tabular}{|l|c|c|c|c|c|}
\hline
$\gamma$ & Top1 & Top5 & Top10 & mAP & SW\\
\hline\hline
$10^{-1}$ & 91.02 & 96.60 & 97.98 & 77.92 & 0.9995\\
$10^{-2}$ & 91.30 & 96.52 & 98.10 & 78.37 & 0.9995\\
\textbf{$10^{-3}$} & \textbf{91.30} & \textbf{96.60} & \textbf{97.70} & \textbf{78.50} & 0.9986\\
$10^{-4}$ & 91.07 & 96.46 & 98.06 & 78.26 & 0.9984\\
0 & 90.95 & 96.78 & 97.94 & 78.05 & 0.0816\\
\hline
\end{tabular}
\end{center}

\end{table}

\textbf{Orthogonal constraints.} According to Eq.~\ref{eq:totalloss}, the hyper parameters $\gamma$ controls the trade-off between the joint angular loss and regularization term. We conducted extensive experiments to show $\gamma$'s influence on embedding layer's orthogonality. The value of $\gamma$ is selected from $\{0, 10^{-1}, 10^{-2}, 10^{-3}, 10^{-4}\}$. As we can see from Tab.~\ref{table:gamma}, our model is insensitive to the value of $\gamma$ from $10^{-4}$ to $10^{-1}$, while the introduction of orthogonality regularization improves the model performance (\eg, comparing $\gamma=10^{-3}$ with $\gamma=0$ ).

\textbf{Effect of joint angular loss.} To show the effect of different components of JAL, we conduct experiments on different variants of JAL on the Market1501, CUHK03 labeled and DukeMTMC-ReID datasets. The results are reported in Tab.~\ref{table:loss}, where C stands for JAL with only classification loss. T stands for JAL with only triplet loss and hard example mining. C+T stands for JAL with combined classification loss and triplet loss. JAL represents joint angular loss without orthogonal regularization, and JAL$_o$ represents JAL with orthogonal regularization.

\begin{table*}[h]\normalsize
\begin{center}
\caption{ The effect of different components of JAL. C stands for JAL with only classification loss. T stands for JAL with only triplet loss and hard example mining. C+T stands for JAL with combined classification loss and triplet loss using $\lambda = 0.2$. JAL represents joint angular loss without orthogonal regularization, and JAL$_o$ represents JAL with orthogonal regularization.} 
\label{table:loss}
\begin{tabular}{|l|c|c|c|c|c|c|c|c|c|c|c|c|}
\hline
  &    & \multicolumn{4}{c|}{Market1501} & \multicolumn{3}{c|}{CUHK03 label} & \multicolumn{4}{c|}{DukeMTMC} \\
\hline

Method & Backbone & Top1 & Top5 & Top10 & mAP & Top1 & Top5 & Top10  & Top1 & Top5 & Top10 & mAP\\
\hline\hline
C & \multirow{5}{*}{\rotatebox[origin=c]{90}{ResNet50}} & 85.18 & 94.30 & 96.11 & 66.36 & 76.83 & 93.26 & 96.37  & 71.97 & 83.95 & 87.59 & 52.93\\
T & & 86.20 & 94.80 & 96.71 & 71.02 & 86.06 & 97.54 & 98.81 & 76.30 & 87.75 & 91.52 & 61.35\\
C+T & & 88.40 & 95.55 & 96.91 & 74.03 & 86.99 & 97.74 & 99.01 & 77.42 & 88.47 & 91.43 & 62.53\\
JAL & & \textbf{90.95} & \textbf{96.78} & \textbf{97.94} & \textbf{78.05} & \textbf{87.43} & \textbf{97.74} & \textbf{98.74}  & \textbf{80.61} & \textbf{90.56} & \textbf{93.31} & \textbf{65.55}\\
JAL$_o$ & &\textbf{91.28} & \textbf{96.55} & \textbf{97.81} & \textbf{78.56} & \textbf{88.80} & \textbf{98.20} & \textbf{99.50} & \textbf{82.54} & \textbf{91.16} & \textbf{93.76} & \textbf{66.85}\\
\hline
\end{tabular}
\end{center}

\end{table*}

Several important observations could be made from Tab.~\ref{table:loss}.
1) Combining classification loss and triplet loss largely improves the performance over using them individually, with around 7.7\% and 3.0\% increases on mAP on Market1501, 10.2\% and 1.0\% increases on Top1 on CUHK03, 9.6\% and 1.2\% increases on mAP on DukeMTMC-reID, respectively. This shows that these two loss functions are complementary in nature. 2) 
Using the proposed homocentric hypersphere embdding, JAL outperforms the baseline C+T by 4\% on mAP on Market1501, 0.4\% on Top1 on CUHK03, and 3\% on mAP on DukeMTMC. This demonstrates the benefit of Joint Angular Loss, which optimizes angle distances rather than Euclidean distances. 3) The orthogonal regularization on weights provides further performance boosts from 0.5\% to 2\% on CMC Top1 and mAP on the three datasets.

\begin{table}\normalsize
\begin{center}
\caption{Test data augmentation influences on Market1501. The model is trained with joint angular loss and ResNet50 as feature extractor.}
\label{table:testaug}
\begin{tabular}{|l|c|c|c|c|}
\hline
Image Num & Top1 & Top5 & Top10 & mAP\\
\hline\hline
Original & 91.28 & 96.55 & 97.81 & 78.56\\
2 & 91.75 & 96.82 & 98.10 & 79.50 \\
4 & 92.40 & 97.10 & 98.37 & 79.90\\
8 & 92.25 & 97.00 & 98.22 & 80.18 \\
12 & 92.34 & 96.91 & 98.22 & 80.22 \\
16 & 92.10 & 97.00 & 98.37 & 80.31 \\
32 & 92.16 & 97.18 & 98.34 & 80.36 \\
\hline
\end{tabular}
\end{center}
\end{table}

  \textbf{Test data augmentation.} Test data augmentation simulates different viewpoints and occlusion effect of original person image. We apply common data augmentation methods, such as random cropping and flipping. The final feature vector of a given image is produced by averaging all features generated by the augmented images and the original one. In Tab.~\ref{table:testaug} , we show the number of augmented images used for producing the final feature and the ReID performances on Market1501. One can see that data augmentation can improve the performance. In the experiments, we report the results of our method with and without data augmentation for more comprehensive comparison with other methods.

\subsection{Comparison with state-of-the-art methods}

\textbf{Market1501.}  Market1501 is currently the largest benchmark dataset for Person ReID, and many methods have been reported on this dataset. We compare the proposed method with most of the state-of-the-arts, including Discriminative Null Space (DNS)~\cite{zhang2016learning}, Gated Siamese Convolutional Neural Network (G-CNN)~\cite{varior2016gated}, Unlabeled Sample Generation GAN (GAN)~\cite{zheng2017unlabeled}, Deep Transfer Learning (DTL)~\cite{geng2016deep}, Joint Learning Multi-Loss (JLML)~\cite{li2017person}, TriNet~\cite{DBLP:journals/corr/HermansBL17}, Deep Context-aware Features (DCF)~\cite{li2017learning}, Spindle network (Spindle)~\cite{zhao2017spindle}, Supervised Smooth Manifold (SSM)~\cite{bai2017scalable}, Point Set Similarity Feature (PSSF)~\cite{zhou2017point},  Deeply Learned Part-Aligned representation (DLPA)~\cite{zhao2017deeply}, and Pose-driven Deep Convolutional (PDC) model~\cite{su2017pose} . The experimental results are shown in Tab.~\ref{table:market1501}.

\begin{figure}[t]
\includegraphics[scale=0.14]{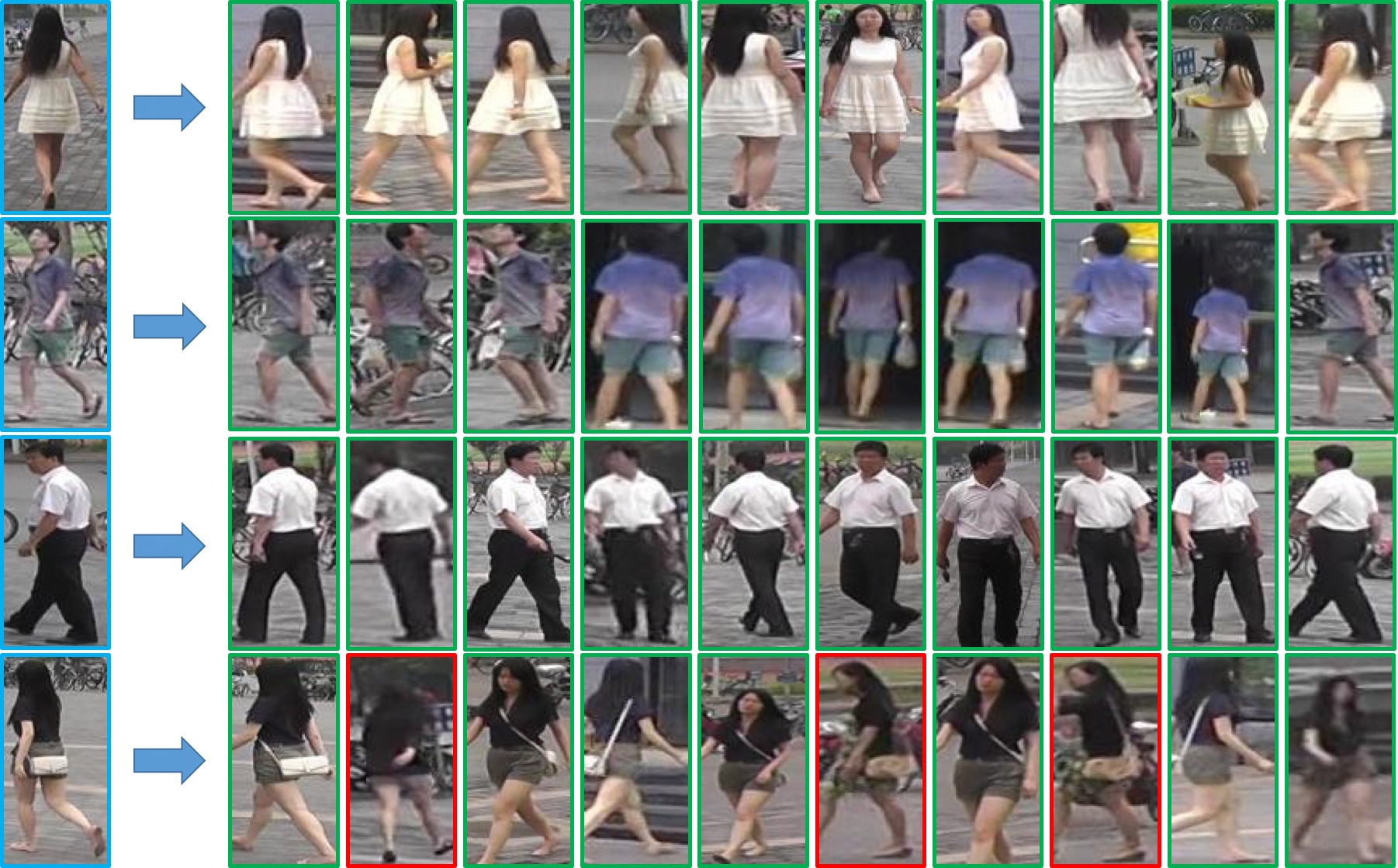}
\centering
\caption{ Sample retrieval results on Market1501 using the proposed method. The images in the first column are the query images. The top-10 retrieved images are sorted according to the similarity scores from left to right. Gallery images captured from the same camera view as query images are already excluded from the ranking list. The correct matches are in the green rectangles, and the false matching images are in the red rectangles. }
\label{fig:query}
\end{figure}

With ResNet50 as the pre-trained network, the proposed JAL approach achieves 78.1\% mAP and 91.0\% CMC top1. It is superior to all compared methods.
 Specifically, JAL outperforms TriNet~\cite{DBLP:journals/corr/HermansBL17} by 6.1\% on CMC top1 and 9\%  on mAP. In addition, JAL also outperforms the newly proposed methods, PLPA and PDC, by a large margin. With the orthogonal regularization and 16 times test augmentation, the performance of JAL is further improved.

In Fig.~\ref{fig:query}, we show the CMC top 10 retrieval results of four query images in Market1501. We can see that JAL exhibits strong robustness to scale, viewpoint and pose. The false matchings marked in red bounding box look very similar to the query image in visual appearance and human pose. These false matchings are even very challenging for human.

\begin{table}\normalsize
\begin{center}
\caption{Single-shot performance comparison of different methods on Market1501. We highlight top-5 results according to mAP.}
\label{table:market1501}
\begin{tabular}{|l|@{}c|@{}c|@{}c|@{}c|}
\hline
Method & Top1 & Top5 & Top10 & mAP\\
\hline\hline
DNS~\cite{zhang2016learning} (CVPR16) & 55.4 & - & - & 29.9\\
G-CNN~\cite{varior2016gated} (ECCV16) & 65.9 & - & - & 39.6\\
GAN~\cite{zheng2017unlabeled} (ICCV17) & 78.1 & - & - & 56.2\\
DTL~\cite{geng2016deep} (Arxiv16) & 83.7 & - & - & 65.5 \\
JLML~\cite{li2017person} (IJCAI17) & 85.1 & - & - & 65.5\\
DCF~\cite{li2017learning} (CVPR17) & 80.3 & - & - & 57.5 \\
Spindle~\cite{zhao2017spindle} (CVPR17) & 76.9 & 91.5 & 94.6 & - \\
PSSF~\cite{zhou2017point} (CVPR17) & 70.7 & 90.5 & - & - \\
DLPA~\cite{zhao2017deeply} (ICCV17) & 81.0 & 92.0 & 94.7 & 63.4\\
PDC~\cite{su2017pose} (ICCV17) & 84.1 & 92.7 & 94.9 & 63.4 \\

SSM~\cite{bai2017scalable} (CVPR17) & \textbf{82.2} & - & - & \textbf{68.8} \\
TriNet~\cite{DBLP:journals/corr/HermansBL17} (Arxiv17) & \textbf{84.9} & \textbf{94.2} & - & \textbf{69.1}\\
\hline\hline
JAL  & \textbf{91.0} & \textbf{96.8} & \textbf{97.9} & \textbf{78.1}\\
JAL$_o$ & \textbf{91.3} & \textbf{96.6} & \textbf{97.8} & \textbf{78.6}\\
JAL$_o$+aug16 & \textbf{92.1} & \textbf{97.0} & \textbf{98.4} & \textbf{80.3} \\
\hline
\end{tabular}
\end{center}
\end{table}

\begin{figure}
\includegraphics[scale=0.16]{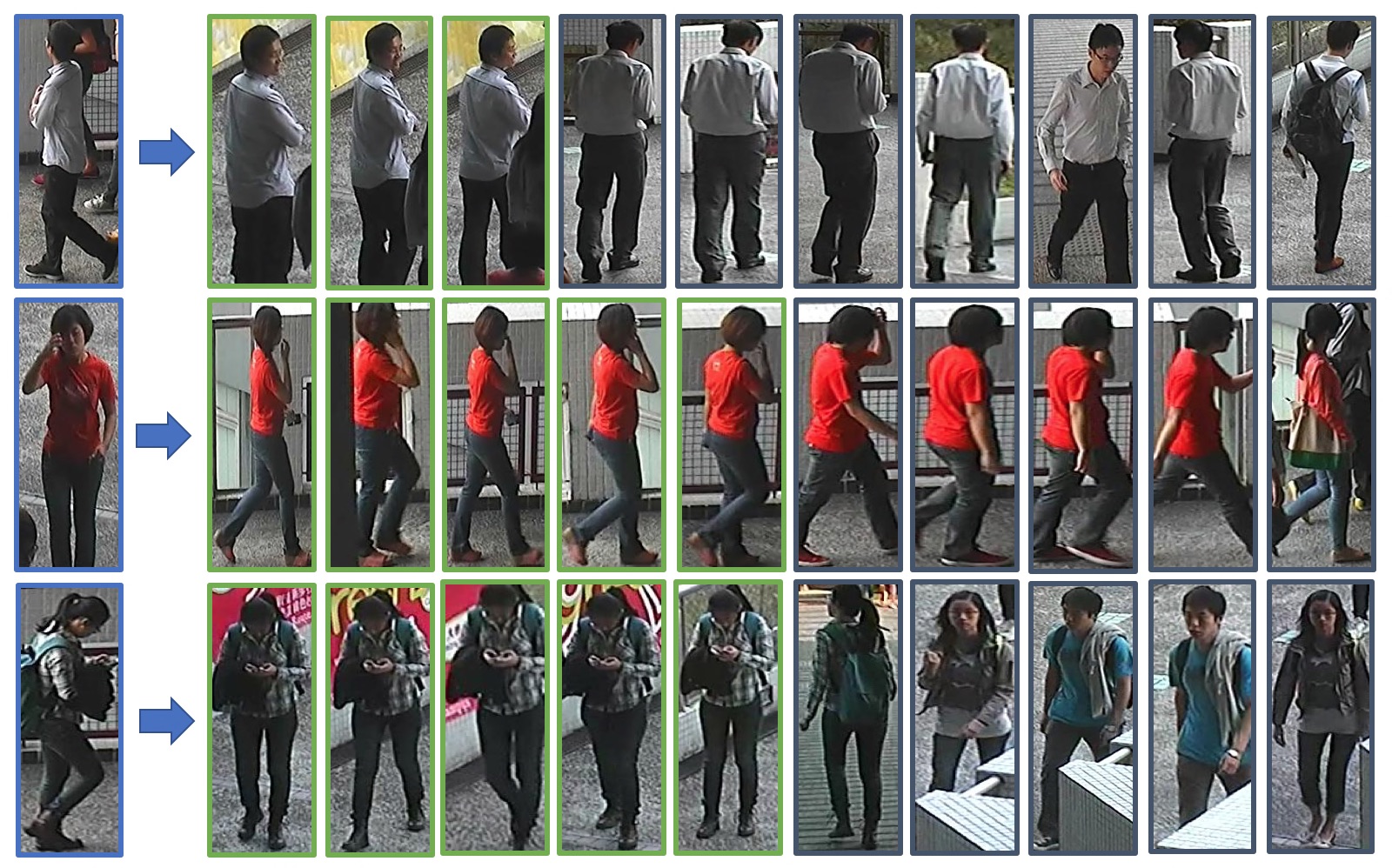}
\centering
\caption{ Sample retrieval results on CUHK03 labeled dataset using the proposed method. The images in the first column are the query images. The top-10 retrieved images are sorted according to the similarity scores from left to right. Gallery images captured from the same camera view as query images are already excluded from the ranking list. The correct matches are in the green rectangles, and the other retrieved images are in the gray rectangles. }
\label{fig:querycuhk03}
\end{figure}

\textbf{CUHK03.} On CUHK03, we follow the standard protocol and conduct experiments using both labeled and detected datasets. We compare our model with Filter Pair Neural Networks (FPNN)~\cite{li2014deepreid}, Improved Deep Learning Architecture (IDLA)~\cite{ahmed2015improved}, Local Maximal Occurrence Representation (LOMO), Sample Specific SVM (SS-SVM)~\cite{zhang2016sample}, Discriminative Null Space (DNS)~\cite{zhang2016learning}, Deep Context-aware Features (DCF), Quadriplet loss (Quadruplet)~\cite{chen2017beyond}, Spindle Network (Spindle), Supervised Smooth Manifold (SSM) ~\cite{bai2017scalable}, Multi-scale Deep Architecture (MuDeep)~\cite{qian2017multi}, Deeply Learned Part-aligned (DLPA)~\cite{zhao2017deeply} and Pose-driven CNN (PDC)~\cite{su2017pose}. The results are shown in Tab.~\ref{table:CUHK03_label} and Tab.~\ref{table:CUHK03_detect}. Fig.~\ref{fig:querycuhk03} shows some sample query results on the CUHK03 labeled dataset. The proposed method retrieves all the images of the same persons from different views and the other retrieved person images are also visually similar with the query samples.

	Using the same hyper parameters as used on the Market1501, our method JAL${_o}$ achieves 88.8\% CMC top1 on CUHK03 labeled dataset and 86.9\% CMC top1 on CUHK03 detected dataset, which are leading results on CUHK03 dataset. It is worth noticing that DLPA~\cite{zhao2017deeply} uses both CUHK03 and Market1501 for training. Spindle~\cite{zhao2017spindle} uses seven data sets for training, including CUHK01, CUHK03, Market1501 and etc. PDC~\cite{su2017pose} fuses two Google Inception sub-networks to consider global feature maps and part feature maps. Our approach is trained on CUHK03 with only a single ResNet50.

\begin{table}\normalsize
\begin{center}
\caption{Single shot performance comparison of different methods on CUHK03 labeled dataset. We hightlight the top-5 performers according to CMC Top1.}
\label{table:CUHK03_label}

\begin{tabular}{|l|c|c|c|c|}
\hline
Method & Top1 & Top5 & Top10 \\
\hline\hline
FPNN~\cite{li2014deepreid} (CVPR14) & 20.7 & 51.5  & 66.5 \\
IDLA~\cite{ahmed2015improved} (CVPR15) & 54.7 & 86.5 & 93.9  \\
LOMO~\cite{7298832} (CVPR15) & 52.2 & 82.2 & 92.1 \\
SS-SVM~\cite{zhang2016sample} (CVPR16) & 57.0 & 84.8 & 92.5  \\
DNS~\cite{zhang2016learning} (CVPR16) & 58.9 & 85.6 & 92.5 \\
DCF~\cite{li2017learning} (CVPR17) & 74.2 & 94.3 & 97.5 \\
Quadruplet~\cite{chen2017beyond} (CVPR17) & 75.5 & 95.2 & 99.2 \\
SSM~\cite{bai2017scalable} (CVPR17) & 76.6 & 94.6 & 98.0  \\
MuDeep~\cite{qian2017multi} (ICCV17) & 76.9 & 96.1 & 98.4 \\
DLPA~\cite{zhao2017deeply} (ICCV17) & 85.4 & 97.6 & 99.4 \\
Spindle~\cite{zhao2017spindle} (CVPR17) & \textbf{88.5} & \textbf{97.8} & \textbf{98.6} \\
PDC~\cite{su2017pose} (ICCV17) & \textbf{88.7} & \textbf{98.6} & \textbf{99.2} \\
\hline\hline

JAL  & \textbf{87.4} & \textbf{97.7} & \textbf{98.7} \\
JAL$_o$ & \textbf{88.8} & \textbf{98.2} & \textbf{99.5} \\
JAL$_o$+aug16 & \textbf{89.5} & \textbf{97.8} & \textbf{99.1} \\

\hline
\end{tabular}
\end{center}
\end{table}

\begin{table}\normalsize
\begin{center}
\caption{Single shot performance comparison of different methods on CUHK03 detected dataset. We hightlight the top-5 performers according to CMC Top1.}
\label{table:CUHK03_detect}

\begin{tabular}{|l|c|c|c|c|}
\hline
Method & Top1 & Top5 & Top10 \\
\hline\hline
FPNN~\cite{li2014deepreid} (CVPR14) & 19.9 & 50.0  & 64.0 \\
IDLA~\cite{ahmed2015improved} (CVPR15) & 45.0 & 76.0 & 83.5  \\
LOMO~\cite{7298832} (CVPR15) & 46.3 & 78.9 & 88.6 \\
SS-SVM~\cite{zhang2016sample} (CVPR16) & 51.2 & 81.5 & 89.9  \\
DNS~\cite{zhang2016learning} (CVPR16) & 54.7 & 84.8 & 94.8 \\
DCF~\cite{li2017learning} (CVPR17) & 68.0 & 91.0 & 95.4 \\
SSM~\cite{bai2017scalable} (CVPR17) & 72.7 & 92.4 & 96.1  \\
MuDeep~\cite{qian2017multi} (ICCV17) & 75.6 & 94.4 & 97.5 \\
PDC~\cite{su2017pose} (ICCV17) & \textbf{78.3} & \textbf{94.8} & \textbf{97.2} \\
DLPA~\cite{zhao2017deeply} (ICCV17) & \textbf{81.6} & \textbf{97.3} & \textbf{98.4} \\

\hline\hline

JAL  & \textbf{84.8} & \textbf{97.0} & \textbf{97.9} \\
JAL$_o$ & \textbf{86.9} & \textbf{97.4} & \textbf{98.5} \\
JAL$_o$+aug16 & \textbf{88.4} & \textbf{97.3} & \textbf{98.4} \\
\hline
\end{tabular}
\end{center}
\end{table}

\begin{figure}
\includegraphics[scale=0.15]{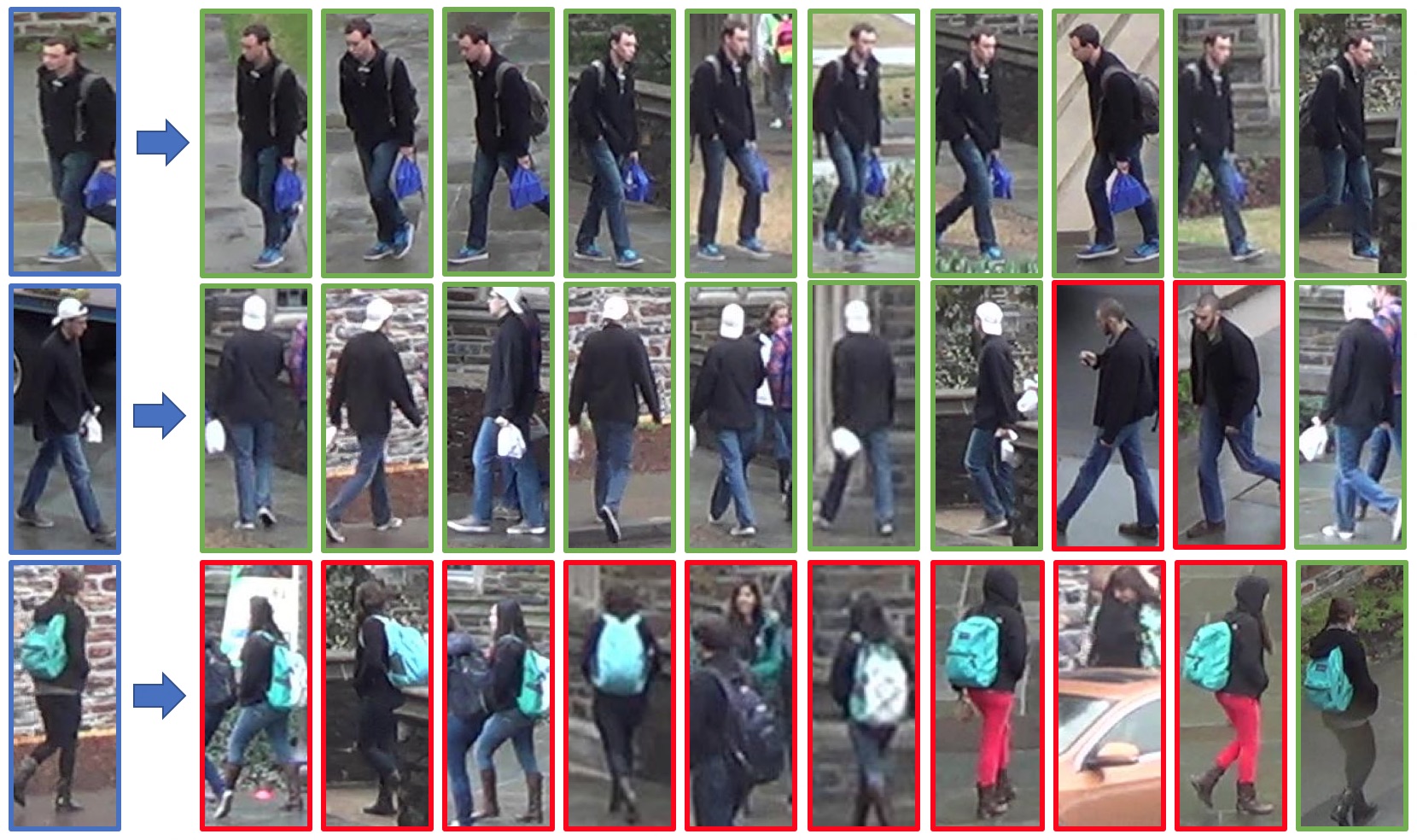}
\centering
\caption{ Retrieval examples on DukeMTMC-ReID dataset using the proposed method. The images in the first column are the query images. The top-10 retrieved images are sorted according to the similarity scores from left to right. Gallery images captured from the same camera view as query images are already excluded from the ranking list. The correct matches are in the green rectangles, and the false matching images are in the red rectangles. }
\label{fig:querydukemtmc}
\end{figure}

\textbf{DukeMTMC-ReID.} On this dataset, we compare our method with Scalable Person Re-identification (BoW+KISSME)~\cite{zheng2015scalable}, Local Maximal Occurrence Representation (LOMO)~\cite{7298832}, Improved Attribute Person ReID (APR)~\cite{DBLP:journals/corr/LinZZWY17}, Unlabeled Sample Generation GAN (GAN), Pedestrian Alignment Network (PAN)~\cite{DBLP:journals/corr/ZhengZY17aa},  SVDNet~\cite{DBLP:journals/corr/SunZDW17} and Deep Learning Multi-Scale Representations (DPFL)~\cite{chen2017person}, which are all state-of-the-art methods we can find in literature.

As shown in Tab.~\ref{table:dukemtmc}, our model achieves much better performance than other methods. The proposed JAL$_o$ obtains 82.5\% Rank-1 accuracy and 66.9\% mAP, respectively. The DPFL~\cite{chen2017person} method fuses multi-scale information by combining multiple sub-networks in the training stage. In contrast, our model is trained only on DukeMTMC-ReID with a single ResNet50. The Fig.~\ref{fig:querydukemtmc} shows some retrieval examples on the DukeMTMC-ReID dataset. In the second row, the person in the wrongly retrieved images has similar clothing to the person in the query image. The third row is a very challenging example, as all the persons wear similar bright green backpacks.

\begin{table}\normalsize
\begin{center}
\caption{Single shot performance comparison of different methods on DukeMTMC-ReID. We highlight the top-5 performers according to mAP. }
\label{table:dukemtmc}
\begin{tabular}{|l|c|c|c|c|}
\hline
Method & Top1 &  mAP\\
\hline\hline
BoW+KISSME~\cite{zheng2015scalable} (CVPR15) & 25.1  & 12.2\\
LOMO~\cite{lomo15liao} (CVPR15) & 30.8  & 17.0 \\
APR~\cite{DBLP:journals/corr/LinZZWY17} (ArXiv17) & 70.7  & 51.9\\
GAN~\cite{zheng2017unlabeled} (ICCV17) & 67.7  & 47.1\\
PAN~\cite{DBLP:journals/corr/ZhengZY17aa} (CVPR17) & 71.6  & 51.5\\
SVDNet~\cite{DBLP:journals/corr/SunZDW17} (ICCV17) & \textbf{76.7}  & \textbf{56.8} \\
DPFL~\cite{chen2017person} (ICCVW17) & \textbf{79.2} & \textbf{60.6} \\
\hline\hline
JAL  & \textbf{80.6} & \textbf{65.6}\\
JAL$_o$ & \textbf{82.5} & \textbf{66.9}\\
JAL$_o$+aug16 & \textbf{83.0} & \textbf{67.8} \\
\hline
\end{tabular}
\end{center}
\end{table}


\section{Conclusion}

In this paper, we proposed a homocentric hypersphere feature embedding learning approach for Person ReID task. In our homecentric hypersphere framework, the class center vectors and the features were normalized to two individual homocentric hyperspheres with the same coordinate origin. Based on the homecentric hypersphere assumption, we reformulated the  classification loss and the triplet loss into their corresponding angular versions, and thus provided a natural way to jointly consider both losses to minimize the angle between intra-class features, maximize the angle between inter-class features and minimize the angle between features and their corresponding class center vectors. To reduce the redundancy in the embedding layers' weights, we placed orthogonal regularizations on embedding layer. Detailed analysis and extensive experiments were conducted on three widely used data sets to validate the effectiveness of our approach for Person ReID. The results showed that our approach achieves state-of-the-art performance on all benchmarks by using the same hyperparameters.

\ifCLASSOPTIONcaptionsoff
  \newpage
\fi



%
%
%

{\small
\bibliographystyle{IEEEtran}
\bibliography{main.bib}

\begin{thebibliography}{10}
\providecommand{\url}[1]{#1}
\csname url@samestyle\endcsname
\providecommand{\newblock}{\relax}
\providecommand{\bibinfo}[2]{#2}
\providecommand{\BIBentrySTDinterwordspacing}{\spaceskip=0pt\relax}
\providecommand{\BIBentryALTinterwordstretchfactor}{4}
\providecommand{\BIBentryALTinterwordspacing}{\spaceskip=\fontdimen2\font plus
\BIBentryALTinterwordstretchfactor\fontdimen3\font minus
  \fontdimen4\font\relax}
\providecommand{\BIBforeignlanguage}[2]{{%
\expandafter\ifx\csname l@#1\endcsname\relax
\typeout{** WARNING: IEEEtran.bst: No hyphenation pattern has been}%
\typeout{** loaded for the language `#1'. Using the pattern for}%
\typeout{** the default language instead.}%
\else
\language=\csname l@#1\endcsname
\fi
#2}}
\providecommand{\BIBdecl}{\relax}
\BIBdecl

\bibitem{gheissari2006person}
N.~Gheissari, T.~B. Sebastian, and R.~Hartley, ``Person reidentification using
  spatiotemporal appearance,'' in \emph{CVPR}, 2006.

\bibitem{7298832}
S.~Liao, Y.~Hu, X.~Zhu, and S.~Z. Li, ``Person re-identification by local
  maximal occurrence representation and metric learning,'' in \emph{CVPR},
  2015.

\bibitem{matsukawa2016hierarchical}
T.~Matsukawa, T.~Okabe, E.~Suzuki, and Y.~Sato, ``Hierarchical gaussian
  descriptor for person re-identification,'' in \emph{CVPR}, 2016.

\bibitem{lecun1995convolutional}
Y.~LeCun, Y.~Bengio \emph{et~al.}, ``Convolutional networks for images, speech,
  and time series,'' \emph{The handbook of brain theory and neural networks},
  1995.

\bibitem{alex}
A.~Krizhevsky, I.~Sutskever, and G.~E. Hinton, ``Imagenet classification with
  deep convolutional neural networks,'' in \emph{NIPS}, 2012.

\bibitem{Simonyan14c}
K.~Simonyan and A.~Zisserman, ``Very deep convolutional networks for
  large-scale image recognition,'' \emph{arXiv preprint arXiv:1409.1556}, 2014.

\bibitem{g1}
C.~Szegedy, W.~Liu, Y.~Jia, P.~Sermanet, S.~Reed, D.~Anguelov, D.~Erhan,
  V.~Vanhoucke, and A.~Rabinovich, ``Going deeper with convolutions,'' in
  \emph{CVPR}, 2015.

\bibitem{He2015}
K.~He, X.~Zhang, S.~Ren, and J.~Sun, ``Deep residual learning for image
  recognition,'' in \emph{CVPR}, 2016.

\bibitem{zheng2015scalable}
L.~Zheng, L.~Shen, L.~Tian, S.~Wang, J.~Wang, and Q.~Tian, ``Scalable person
  re-identification: A benchmark,'' in \emph{CVPR}, 2015.

\bibitem{li2012human}
W.~Li, R.~Zhao, and X.~Wang, ``Human reidentification with transferred metric
  learning,'' in \emph{ACCV}, 2012.

\bibitem{li2014deepreid}
W.~Li, R.~Zhao, T.~Xiao, and X.~Wang, ``Deepreid: Deep filter pairing neural
  network for person re-identification,'' in \emph{CVPR}, 2014.

\bibitem{ristani2016MTMC}
E.~Ristani, F.~Solera, R.~Zou, R.~Cucchiara, and C.~Tomasi, ``Performance
  measures and a data set for multi-target, multi-camera tracking,'' in
  \emph{ECCV workshop}, 2016.

\bibitem{chen2017beyond}
W.~Chen, X.~Chen, J.~Zhang, and K.~Huang, ``Beyond triplet loss: a deep
  quadruplet network for person re-identification,'' in \emph{CVPR}, 2017.

\bibitem{DBLP:journals/corr/SunZDW17}
Y.~Sun, L.~Zheng, W.~Deng, and S.~Wang, ``Svdnet for pedestrian retrieval,'' in
  \emph{ICCV}, 2017.

\bibitem{zhao2017deeply}
L.~Zhao, X.~Li, J.~Wang, and Y.~Zhuang, ``Deeply-learned part-aligned
  representations for person re-identification,'' in \emph{ICCV}, 2017.

\bibitem{varior2016gated}
R.~R. Varior, M.~Haloi, and G.~Wang, ``Gated siamese convolutional neural
  network architecture for human re-identification,'' in \emph{ECCV}, 2016.

\bibitem{DBLP:journals/corr/VariorSLXW16}
R.~R. Varior, B.~Shuai, J.~Lu, D.~Xu, and G.~Wang, ``A siamese long short-term
  memory architecture for human re-identification,'' in \emph{ECCV}, 2016.

\bibitem{geng2016deep}
M.~Geng, Y.~Wang, T.~Xiang, and Y.~Tian, ``Deep transfer learning for person
  re-identification,'' \emph{arXiv preprint arXiv:1611.05244}, 2016.

\bibitem{DBLP:journals/corr/YiLL14}
D.~Yi, Z.~Lei, and S.~Z. Li, ``Deep metric learning for practical person
  re-identification,'' \emph{arXiv preprint arXiv:1407.4979}, 2014.

\bibitem{DBLP:journals/corr/ShiYZLLZL16}
H.~Shi, Y.~Yang, X.~Zhu, S.~Liao, Z.~Lei, W.~Zheng, and S.~Z. Li, ``Embedding
  deep metric for person re-identication {A} study against large variations,''
  in \emph{ECCV}, 2016.

\bibitem{jose2016scalable}
C.~Jose and F.~Fleuret, ``Scalable metric learning via weighted approximate
  rank component analysis,'' in \emph{ECCV}, 2016.

\bibitem{DBLP:journals/corr/SongXJS15}
H.~O. Song, Y.~Xiang, S.~Jegelka, and S.~Savarese, ``Deep metric learning via
  lifted structured feature embedding,'' in \emph{CVPR}, 2016.

\bibitem{liao2015efficient}
S.~Liao and S.~Z. Li, ``Efficient psd constrained asymmetric metric learning
  for person re-identification,'' in \emph{ICCV}, 2015.

\bibitem{li2017learning}
D.~Li, X.~Chen, Z.~Zhang, and K.~Huang, ``Learning deep context-aware features
  over body and latent parts for person re-identification,'' in \emph{CVPR},
  2017.

\bibitem{zhao2017spindle}
H.~Zhao, M.~Tian, S.~Sun, J.~Shao, J.~Yan, S.~Yi, X.~Wang, and X.~Tang,
  ``Spindle net: Person re-identification with human body region guided feature
  decomposition and fusion,'' in \emph{CVPR}, 2017.

\bibitem{DBLP:journals/corr/ZhengZY17aa}
Z.~Zheng, L.~Zheng, and Y.~Yang, ``Pedestrian alignment network for large-scale
  person re-identification,'' in \emph{CVPR}, 2017.

\bibitem{zhang2016semantics}
Y.~Zhang, X.~Li, L.~Zhao, and Z.~Zhang, ``Semantics-aware deep correspondence
  structure learning for robust person re-identification.'' in \emph{IJCAI},
  2016.

\bibitem{DBLP:journals/corr/HermansBL17}
A.~Hermans, L.~Beyer, and B.~Leibe, ``In defense of the triplet loss for person
  re-identification,'' \emph{arXiv preprint arXiv:1703.07737}, 2017.

\bibitem{cheng2016person}
D.~Cheng, Y.~Gong, S.~Zhou, J.~Wang, and N.~Zheng, ``Person re-identification
  by multi-channel parts-based cnn with improved triplet loss function,'' in
  \emph{CVPR}, 2016.

\bibitem{li2017person}
W.~Li, X.~Zhu, and S.~Gong, ``Person re-identification by deep joint learning
  of multi-loss classification,'' in \emph{IJCAI}, 2017.

\bibitem{xiao2016learning}
T.~Xiao, H.~Li, W.~Ouyang, and X.~Wang, ``Learning deep feature representations
  with domain guided dropout for person re-identification,'' in \emph{CVPR},
  2016.

\bibitem{DeCov}
M.~Cogswell, F.~Ahmed, R.~B. Girshick, L.~Zitnick, and D.~Batra, ``Reducing
  overfitting in deep networks by decorrelating representations,'' in
  \emph{ICLR}, 2016.

\bibitem{PCANet}
T.~H. Chan, K.~Jia, S.~Gao, J.~Lu, Z.~Zeng, and Y.~Ma, ``Pcanet: A simple deep
  learning baseline for image classification?'' in \emph{IEEE Transactions on
  Image Processing}, vol.~24, no.~12, Dec 2015, pp. 5017--5032.

\bibitem{xie2017all}
D.~Xie, J.~Xiong, and S.~Pu, ``All you need is beyond a good init: Exploring
  better solution for training extremely deep convolutional neural networks
  with orthonormality and modulation,'' in \emph{CVPR}, 2017.

\bibitem{chen2017person}
Y.~Chen, X.~Zhu, and S.~Gong, ``Person re-identification by deep learning
  multi-scale representations,'' in \emph{CVPR}, 2017.

\bibitem{zhou2017large}
S.~Zhou, J.~Wang, R.~Shi, Q.~Hou, Y.~Gong, and N.~Zheng, ``Large margin
  learning in set to set similarity comparison for person re-identification,''
  \emph{IEEE Transactions on Multimedia}, 2017.

\bibitem{ahmed2015improved}
E.~Ahmed, M.~Jones, and T.~K. Marks, ``An improved deep learning architecture
  for person re-identification,'' in \emph{CVPR}, 2015.

\bibitem{wu2016personnet}
L.~Wu, C.~Shen, and A.~v.~d. Hengel, ``Personnet: Person re-identification with
  deep convolutional neural networks,'' \emph{arXiv preprint arXiv:1601.07255},
  2016.

\bibitem{zheng2016person}
L.~Zheng, Y.~Yang, and A.~G. Hauptmann, ``Person re-identification: Past,
  present and future,'' \emph{arXiv preprint arXiv:1610.02984}, 2016.

\bibitem{liu2017sphereface}
W.~Liu, Y.~Wen, Z.~Yu, M.~Li, B.~Raj, and L.~Song, ``Sphereface: Deep
  hypersphere embedding for face recognition,'' in \emph{CVPR}, 2017.

\bibitem{liu2017rethinking}
Y.~Liu, H.~Li, and X.~Wang, ``Rethinking feature discrimination and
  polymerization for large-scale recognition,'' in \emph{NIPS 2017 workshop},
  2017.

\bibitem{Salimans2016WeightNorm}
T.~Salimans and D.~P. Kingma, ``Weight normalization: A simple
  reparameterization to accelerate training of deep neural networks,'' in
  \emph{NIPS}, 2016.

\bibitem{schroff2015facenet}
F.~Schroff, D.~Kalenichenko, and J.~Philbin, ``Facenet: A unified embedding for
  face recognition and clustering,'' in \emph{CVPR}, 2015.

\bibitem{weinberger2006distance}
K.~Q. Weinberger, J.~Blitzer, and L.~K. Saul, ``Distance metric learning for
  large margin nearest neighbor classification,'' in \emph{NIPS}, 2006.

\bibitem{bishop2006pattern}
C.~M. Bishop, \emph{Pattern recognition and machine learning}, 2006.

\bibitem{deng2009imagenet}
J.~Deng, W.~Dong, R.~Socher, L.-J. Li, K.~Li, and L.~Fei-Fei, ``Imagenet: A
  large-scale hierarchical image database,'' in \emph{CVPR}, 2009.

\bibitem{felzenszwalb2010object}
P.~F. Felzenszwalb, R.~B. Girshick, D.~McAllester, and D.~Ramanan, ``Object
  detection with discriminatively trained part-based models,'' \emph{TPAMI},
  2010.

\bibitem{zheng2017unlabeled}
Z.~Zheng, L.~Zheng, and Y.~Yang, ``Unlabeled samples generated by gan improve
  the person re-identification baseline in vitro,'' in \emph{CVPR}, 2017.

\bibitem{randomerase}
G.~K. S. L. Y.~Y. Zhun~Zhong, Liang~Zheng, ``Random erasing data
  augmentation,'' \emph{arXiv preprint arXiv:1708.04896}, 2017.

\bibitem{zhang2016learning}
L.~Zhang, T.~Xiang, and S.~Gong, ``Learning a discriminative null space for
  person re-identification,'' in \emph{CVPR}, 2016.

\bibitem{bai2017scalable}
S.~Bai, X.~Bai, and Q.~Tian, ``Scalable person re-identification on supervised
  smoothed manifold,'' 2017.

\bibitem{zhou2017point}
S.~Zhou, J.~Wang, J.~Wang, Y.~Gong, and N.~Zheng, ``Point to set similarity
  based deep feature learning for person re-identification,'' in \emph{CVPR},
  2017.

\bibitem{su2017pose}
C.~Su, J.~Li, S.~Zhang, J.~Xing, W.~Gao, and Q.~Tian, ``Pose-driven deep
  convolutional model for person re-identification,'' in \emph{CVPR}, 2017.

\bibitem{zhang2016sample}
Y.~Zhang, B.~Li, H.~Lu, A.~Irie, and X.~Ruan, ``Sample-specific svm learning
  for person re-identification,'' in \emph{CVPR}, 2016.

\bibitem{qian2017multi}
X.~Qian, Y.~Fu, Y.-G. Jiang, T.~Xiang, and X.~Xue, ``Multi-scale deep learning
  architectures for person re-identification,'' in \emph{ICCV}, 2017.

\bibitem{DBLP:journals/corr/LinZZWY17}
Y.~Lin, L.~Zheng, Z.~Zheng, Y.~Wu, and Y.~Yang, ``Improving person
  re-identification by attribute and identity learning,'' \emph{arXiv preprint
  arXiv:1703.07220}, 2017.

\bibitem{lomo15liao}
X.~Z. Shengcai~Liao, Yang~Hu and S.~Z. Li, ``Person re-identification by local
  maximal occurrence representation and metric learning,'' in \emph{CVPR},
  2015.

\end{thebibliography}
}

%









\end{document}